\documentclass[letterpaper]{article}
\usepackage{aaai2027}
\usepackage[hyphens]{url}
\usepackage{graphicx}
\usepackage{natbib}
\usepackage{caption}
\usepackage{booktabs}

\usepackage{amsmath,amsfonts,bm}

\def\eqref#1{equation~\ref{#1}}

\def\1{\bm{1}}

\DeclareMathAlphabet{\mathsfit}{\encodingdefault}{\sfdefault}{m}{sl}
\SetMathAlphabet{\mathsfit}{bold}{\encodingdefault}{\sfdefault}{bx}{n}

\usepackage{microtype}

\usepackage{pifont}
\usepackage{dsfont}
\usepackage{bm}
\usepackage{nicefrac}
\usepackage{xspace}

\usepackage{amsmath}
\usepackage{amssymb}
\usepackage{mathtools}
\usepackage{amsthm}
\usepackage{thmtools}
\usepackage{amsfonts}
\usepackage{siunitx}

\usepackage{booktabs}
\usepackage{tabularx}
\usepackage{array}
\usepackage{multirow}
\usepackage{colortbl}
\usepackage{makecell}
\usepackage{pbox}
\usepackage{longtable}
\usepackage{rotating}
\usepackage{tablefootnote}
\usepackage{bigstrut}

\usepackage{graphicx}
\usepackage{xcolor}
\usepackage{subcaption}
\usepackage[most]{tcolorbox}
\usepackage{fontawesome5}
\usepackage[colorlinks=true,urlcolor=black,citecolor=black,linkcolor=black]{hyperref}

\newcommand{\squishlist}{
   \begin{list}{$\bullet$}
    { \setlength{\itemsep}{0pt}      \setlength{\parsep}{3pt}
      \setlength{\topsep}{3pt}       \setlength{\partopsep}{0pt}
      \setlength{\leftmargin}{1.0em} \setlength{\labelwidth}{1em}
      \setlength{\labelsep}{0.5em} } }
      
\newcommand{\squishend}{
    \end{list}  }

\definecolor{my_green}{rgb}{0.0, 0.6, 0.0}

\definecolor{codegreen}{rgb}{0,0.6,0}
\definecolor{codegray}{rgb}{0.5,0.5,0.5}
\definecolor{codepurple}{rgb}{0.58,0,0.82}
\definecolor{backcolour}{rgb}{0.95,0.95,0.92}
\definecolor{codeblue}{RGB}{0,0,128} 
\definecolor{blanchedalmond}{rgb}{1.0, 0.92, 0.8}
\definecolor{carmine}{rgb}{0.59, 0.0, 0.09}
\definecolor{amaranth}{rgb}{0.9, 0.17, 0.31}
\definecolor{antiquebrass}{rgb}{0.8, 0.58, 0.46}
\definecolor{antiquefuchsia}{rgb}{0.57, 0.36, 0.51}
\definecolor{chromeyellow}{rgb}{0.31, 0.47, 0.26}

\definecolor{lightblue}{rgb}{0.22,0.45,0.70}%
\definecolor{Gray}{gray}{0.95}
\definecolor{Cornsilk}{rgb}{1.0, 0.97, 0.86}
\definecolor{lightyellow}{RGB}{255,255,204}
\definecolor{lavender}{RGB}{230,220,245}
\definecolor{mintgreen}{RGB}{200,230,210}
\definecolor{softblue}{RGB}{200,220,245}

\newtcolorbox{promptbox}[2][]{%
    enhanced,
    colback=gray!4!white,
    colframe=gray!60!black,
    colbacktitle=gray!75!black,
    coltitle=white,
    fonttitle=\bfseries\sffamily\small,
    title={#2},
    arc=3pt,
    boxrule=0.6pt,
    left=8pt, right=8pt, top=6pt, bottom=6pt,
    fontupper=\small\ttfamily,
    #1
}

\title{The Low-Frequency Trap: Video--Language Models Fail at Simple Event Bookkeeping}
\author{
    Sarvesh Baskar\textsuperscript{1}\equalcontrib, Zikui Cai\textsuperscript{1}\equalcontrib, Shayan Shabihi\textsuperscript{1}\equalcontrib,
    Anirudh Satheesh\textsuperscript{1}, Muhammad R.~Islam\textsuperscript{1}, \\ Udari Madhushani Sehwag\textsuperscript{2},
    Tom Goldstein\textsuperscript{1}, Furong Huang\textsuperscript{1}
}
\affiliations{
    \textsuperscript{1}University of Maryland, College Park, USA \quad
    \textsuperscript{2}Scale AI, USA\\[4pt]
    \small
    \textcolor[HTML]{24292E}{\faGithub}~\textbf{Code:}~\href{https://github.com/Low-Frequency-Trap/The-Low-Frequency-Trap}{\textcolor[HTML]{2563EB}{\texttt{The-Low-Frequency-Trap}}} \quad
    \textcolor[HTML]{D97706}{\faDatabase}~\textbf{Dataset:}~\href{https://huggingface.co/datasets/Sarvesh-369/Low-Frequency-Trap}{\textcolor[HTML]{D97706}{\texttt{Sarvesh-369/Low-Frequency-Trap}}}\\[2pt]
    \textcolor[HTML]{0284C7}{\faGlobe}~\textbf{Project Page:}~\href{https://low-frequency-trap.github.io}{\textcolor[HTML]{0284C7}{\texttt{low-frequency-trap.github.io}}}
}

\begin{document}

\maketitle

\begin{abstract}
Real-world video benchmarks provide broad coverage, but their fixed clips entangle event count, rate, duration, and visual complexity, making failure modes hard to isolate. While existing programmatic benchmarks offer better control, they primarily score only the final answer rather than auditing reported events against executable ground truth. To bridge this gap, we introduce trace-grounded parametric profiling for event counting in three controlled video tasks, bouncing-ball wall contacts, visual blinks, and categorical state transitions. Across 2,190 videos, we vary event count \(N\) and frequency \(F\) while holding rendering fixed. Each video includes an executable event trace for capability-surface estimation and timestamp-level evaluation. Our results reveal a staged temporal failure. At an 80\% reliability threshold, Gemini 3.6 Flash reliably counts persistent state transitions up to 12 events at 0.5 and 1.0 Hz, yet demonstrates no reliable positive-count region for transient blinking events. Thus, event representation (persistent or transient) dictates whether a model initially accesses the evidence -- a limitation that compounds as count and frequency increase. In the high-count, high-frequency regime, only 0.2\% of final counts are correct and the model recovers just 18.1\% of true events. To test if visual access is the primary bottleneck, we increase sampling rate. Although this boosts Bounce Ball accuracy from 19.6\% to 29.3\%, the reported sequence agrees with the ground truth only 3.7\% of the time. Extra frames can therefore inflate final scores without producing faithful event recovery. Different prompting strategies yield similarly limited gains, and evaluations on real-world videos show the same concentration of success at low event counts. Ultimately, trace-grounded profiling shifts video evaluation from aggregate accuracy metrics to a detailed diagnostic of where temporal reasoning fails and whether the reported evidence actually supports the final answer.
\end{abstract}

\section{Introduction}
\label{sec:introduction}

\begin{figure*}[t]
    \centering
    \includegraphics[width=0.9\textwidth]{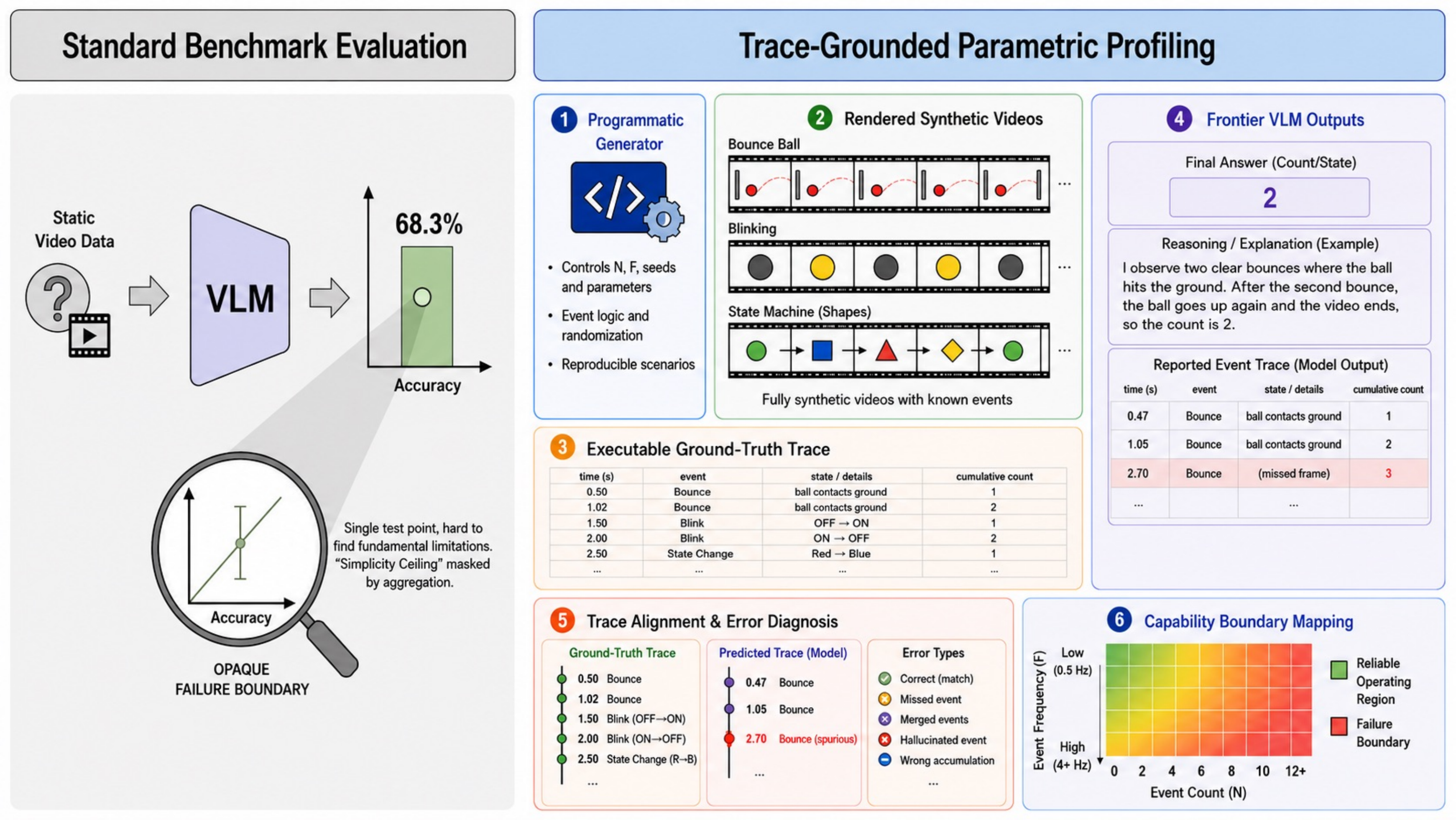}
    \caption{Overview of trace-grounded parametric profiling. Unlike standard benchmark evaluation based on a single aggregate accuracy score, our framework generates controlled videos with executable event traces.
}
    \label{fig:overview_parametric_profiling}
\end{figure*}

Video-language models have advanced rapidly, demonstrating strong performance across standard video benchmarks \citep{fu2025videomme, liu2024tempcompass, chandrasegaran2024hourvideo, mangalam2023egoschemadiagnosticbenchmarklongform, li2024mvbench}. These benchmarks provide useful breadth and ranking leaderboards, but they rarely identify why a model fails. Natural clips jointly vary visual clutter, camera motion, event density, duration, and semantics, so an error cannot be attributed to a specific temporal demand. The programmatically synthesized benchmarks considered here offer more control \citep{videoniah, videocogqa}, but primarily score only a final answer. They therefore do not reveal whether a model missed events, confused their timing, lost track of a sequence, or made an error only when aggregating it.

We address this gap with trace-grounded parametric profiling for visual event counting (Figure~\ref{fig:overview_parametric_profiling}). We construct simple controlled video tasks based on bouncing-ball wall contacts, visual blinks, and categorical state transitions. We vary event count $N$ because it determines the number of observations a model must retain and aggregate, and event frequency $F$ because it determines how closely successive events are separated in time. These two parameters capture complementary temporal demands while remaining independently controllable. We hold rendering, semantics, and task rules fixed so that the resulting boundary can be attributed to temporal demand rather than changes in visual content. Other factors that shape real-world video understanding, including clutter, camera motion, occlusion, semantic complexity, and temporal parameters such as event duration, irregularity, and inter-event variability, are outside the scope of this controlled study. The natural-video transfer evaluation does not isolate these factors, it tests only whether the failure patterns identified under controlled variations of $N$ and $F$ persist in more realistic videos. Every video is paired with an executable ground-truth trace containing event times, state changes, and cumulative counts. This design maps reliable operating regions over the $N \times F$ space and tests model-reported events against the sequence that actually occurred.

The results are consistent with a staged and representation-dependent temporal failure. Under an $80\%$ reliability target, Gemini 3.6 Flash counts persistent state transitions through $N=12$ at $0.5$ and $1.0$ Hz, whereas blinking has no reliable positive-count region. In the high-count, high-frequency region, only $0.2\%$ of final counts are correct and Gemini recovers just $18.1\%$ of the true events. Increasing frame density raises Bounce Ball accuracy from $19.6\%$ to $29.3\%$, yet the reported sequence agrees with the ground truth only $3.7\%$ of the time. Thus, additional visual evidence can improve the final answer without producing faithful event recovery. Prompting yields similarly limited gains, and natural repeated-event videos show the same concentration of success at low event counts. Together, these findings are consistent with a staged failure involving event access followed by sequence retention and aggregation. Our paper makes four primary contributions.
\begin{itemize}
    \item We introduce a controlled evaluation framework that replaces a single aggregate score with capability surfaces over event count and frequency.
    \item We provide a trace-grounded benchmark that audits model-reported events against executable ground truth.
    \item We conduct targeted interventions using denser sampling, event-centered keyframes, and prompt variants to diagnose whether errors arise from limited visual evidence, event localization, or reasoning over the recovered sequence.
    \item We show why final-answer accuracy alone can mischaracterize temporal reasoning and provide evidence that the controlled failure pattern extends to natural repeated-event videos.
\end{itemize}

\section{Related Work}
\label{sec:related_work}

\paragraph{Video-language evaluation.}
Recent video-language models combine strong visual encoders with long-context reasoning, including Qwen, InternVL, and Molmo families \cite{bai2025qwen3vltechnicalreport,wang2025internvl35advancingopensourcemultimodal,clark2026molmo2openweightsdata}. Existing benchmarks assess broad temporal understanding on fixed, heterogeneous clips, including Video-MME, TempCompass, HourVideo, EgoSchema, MVBench, LongVideoBench, Mementos, VideoNIAH, VideoCogQA, Video-MMLU, VideoVista, CinePile, Video-MMMU, and V-STaR \cite{fu2025videomme,liu2024tempcompass,chandrasegaran2024hourvideo,mangalam2023egoschemadiagnosticbenchmarklongform,li2024mvbench,wu2024longvideobenchbenchmarklongcontextinterleaved,wang2024mementos,videoniah,videocogqa,song2025videommlumassivemultidisciplinelecture,li2024videovista,rawal2024cinepile,hu2025video,cheng2026vstar}. VideoReasonBench additionally tests multi-step reasoning over videos with partially observed state changes \cite{liu2025videoreasonbench}. Studies of visual shortcuts, temporal perturbations, and benchmark confounds likewise show that final-answer scores can conceal weak temporal access or brittle reasoning \cite{fu2024blinkmultimodallargelanguage,tong2024eyeswideshutexploring,cores2025lostintime,feng2025breaking,yue2024mmmumassivemultidisciplinemultimodal,lu2024mathvistaevaluatingmathematicalreasoning,chen2024mmstar,hao2025emma}. Our framework complements this literature by holding rendering and semantics fixed while sweeping explicit event-load and event-rate demands.

\paragraph{Controlled diagnostic evaluation.}
Controlled visual benchmarks, from CLEVR to recent physical-reasoning probes, isolate factors that are entangled in natural data \cite{johnson2017clevr,chow2025physbench,xiang2025seephys,qiu2025phybench}. Complementary analyses such as RCI ask whether a benchmark genuinely requires global visual reasoning \cite{agarwal2025rciscoreevaluatingglobal}. We use the same diagnostic principle for video. Programmatic generation gives exact event boundaries and permits matched interventions, while the natural-video study tests whether the resulting diagnosis transfers beyond the controlled setting.

\paragraph{Grounded video evaluation.}
MORSE-500 demonstrates the value of fully scripted, controllable videos for stress-testing multimodal reasoning \cite{cai2025morse500programmaticallycontrollablevideo}. TransRAC studies repetitive-action counting from natural video \cite{hu2022transrac}, but does not evaluate model-reported temporal traces. The closest related trace evaluation we found is Visual Reasoning Tracer, which evaluates object-level intermediate reasoning traces in images \cite{yuan2025visual}. Our setting differs because the renderer produces a temporal event schedule. We align a model-reported event sequence with that schedule to distinguish unsupported correct counts from faithful event recovery. Appendix~\ref{app:extended_related_work} provides a detailed comparison with controlled video benchmarks and intermediate-trace evaluations.

\section{Controlled Profiling and Trace Evaluation}
\label{sec:profiling_framework}
\begin{figure*}[t]
    \centering
    \includegraphics[width=0.9\textwidth]{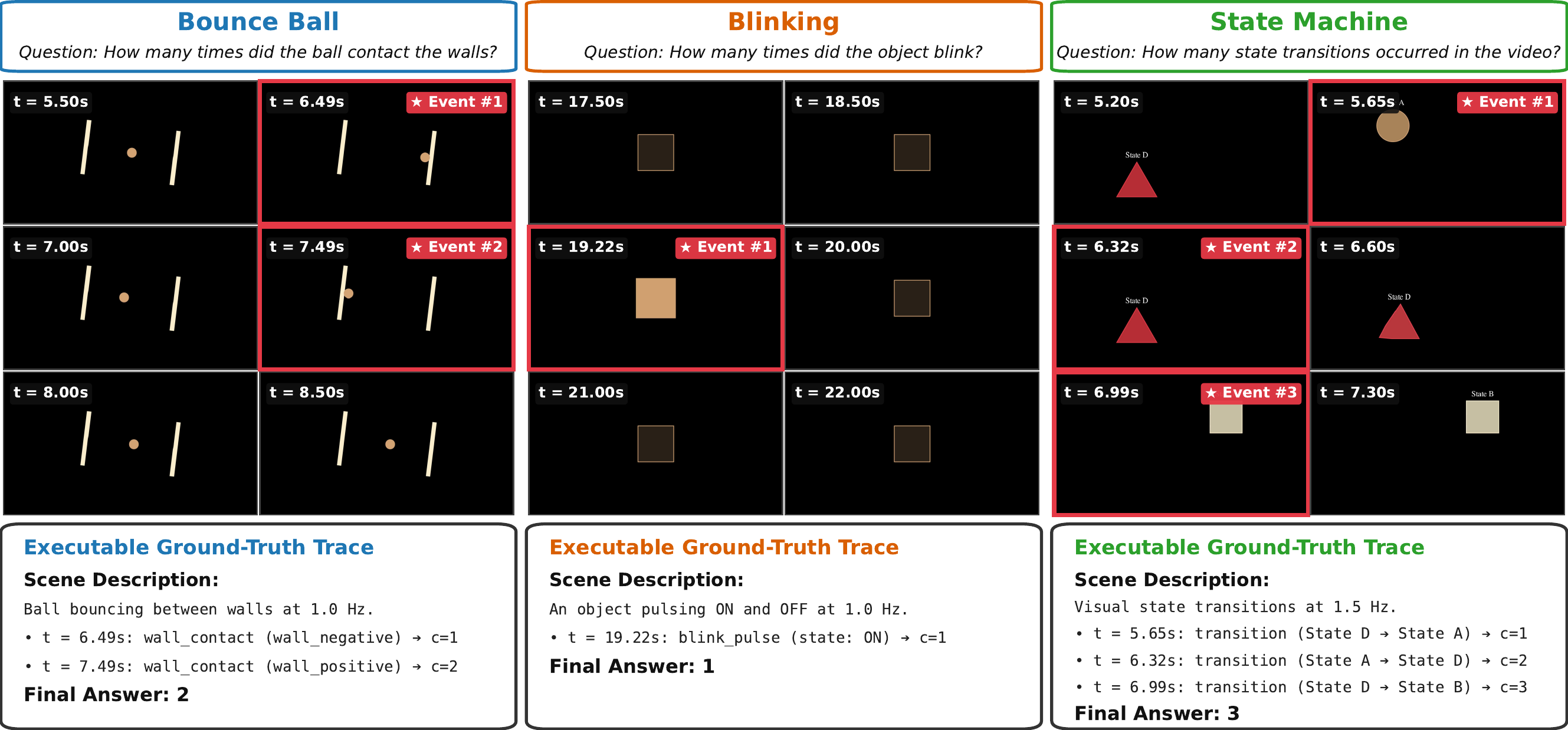}
    \caption{\textbf{Controlled videos and executable traces.} For each \((N,F)\), the renderer records event times, state changes, and cumulative count \(c_i\) for alignment with model-reported timestamps.}
    \label{fig:data_overview}
\end{figure*}

Real-world video benchmarks entangle event count, event rate,
duration, visual complexity, and semantic content, making it
difficult to isolate the source of a model failure. We address
this limitation with programmatically generated videos whose
event structure is known exactly. Each video is paired with an
executable trace produced by the same event schedule used for
rendering. Appendix~\ref{app:synthetic_generation_details} provides
the complete task specifications and supplied-frame examples.

\paragraph{Model evaluation setup.}
We select Gemini 3.6 Flash as the primary deployed video-language
model because it is a frontier model that supports native video input. We use Qwen3-VL-235B, a
large open-weight model from a distinct family, as a supplementary
cross-system check. This tests qualitative robustness, not
universality across VLMs. Each model uses its native video interface
and the same zero-shot template
instantiated for each domain, without per-cell tuning. Input
configurations provide Gemini with 1 FPS and Qwen with 2 FPS; these
are supplied sampling rates, not undocumented internal processing.
We retain default provider decoding and preprocessing. Full prompts
are provided in Appendix~\ref{app:prompting_details}. Detailed
model configurations and cross-model boundary summaries appear in
Appendix~\ref{app:cross_model_results}.

\subsection{Controlled Temporal Task Space}
\label{subsec:parametric_profiling}

Each video is defined by a domain \(d\), event count \(N\), event
frequency \(F\), and random seed \(s\). Figure~\ref{fig:data_overview}
defines an event as a wall contact in \texttt{Bounce\_ball}, an
on-pulse in \texttt{Blinking}, or a categorical transition in
\texttt{State\_machine}. The seed varies color, geometry,
trajectory, initial state, orientation, and start delay while preserving the
event schedule.

We evaluate
\[
\mathcal{N}=\{0,1,2,3,4,5,6,8,10,12\}
\]
and
\[
\mathcal{F}=\{0.5,1.0,1.5,2.0,2.5,3.0,3.5,4.0\}\ \text{Hz}
\]

Every clip is 24 seconds long. The active sequence spans
approximately \(N/F\), the remaining static frames occur before or
after it, with allocation randomized by \(s\). Thus, \(N\) changes
event load and active retention span, not total duration. To balance
grid coverage and visual variation, we generate \(S=10\) videos per
\((N,F)\). Let
\(\hat y_{d,N,F,s}\) be the prediction of model
\(\mathcal{M}_{\theta}\). The observed cell accuracy is

\begin{equation}
\widehat{\mathcal{P}}_{\theta}^{d}(N,F)
=
\frac{1}{S}
\sum_{s=1}^{S}
\mathbb{I}
\left[
\hat y_{d,N,F,s}=N
\right]
\label{eq:parametric_profiling}
\end{equation}

where \(\mathbb{I}[\cdot]\) equals one when the prediction is
correct and zero otherwise. The full evaluation contains
\((1+9\times8)\times10=730\) videos per domain and \(2190\)
videos across all three domains.

For \(N=0\), frequency is inapplicable, so we render ten no-event
controls at nominal \(F=1.0\) Hz, they are excluded from trace
ratios. The profile therefore measures event load and active
retention span, rather than isolated arithmetic counting.

\subsection{Executable Traces and Metrics}
\label{subsec:trace_evaluation}

For a video containing \(N\) events, the renderer produces

\begin{equation}
\mathcal{T}^{*}
=
\left(
t_i,e_i,s_i^{-},s_i^{+},c_i
\right)_{i=1}^{N},
\thinspace
c_i=c_{i-1}+1,\thinspace c_0=0,
\label{eq:ground_truth_trace}
\end{equation}

where \(t_i\) is the event timestamp, \(e_i\) is the event type,
\(s_i^{-}\) and \(s_i^{+}\) are the states before and after the
event, and \(c_i\) is the cumulative count. The trace is generated
before inference and is not supplied to the model in the baseline
condition.

When trace reporting is requested, the model returns \(M\)
timestamped events and a final count \(\hat y\), which we parse
before scoring. Events are aligned one-to-one within the
rate-relative window \(\delta(F)=1/(2F)\), half the inter-event
period. It ranges from \(1.0\,\mathrm{s}\) at \(0.5\) Hz to
\(0.125\,\mathrm{s}\) at \(4.0\) Hz, so adjacent windows do not
overlap. Let \(K\) be the aligned pairs. Prompting, matching, and
fixed-window sensitivity results are in
Appendix~\ref{app:trace_evaluation_and_validation}.

For one video, \(N\) is the true number of events, \(M\) is the
number reported by the model, \(K\) is the number of aligned event
pairs, and \(\hat y\) is the reported final count.
\(\mathbb{I}[\cdot]\) equals one when its condition holds and zero
otherwise. The main evaluation metrics are

\begin{align}
\mathrm{Acc}
&=
\mathbb{I}[\hat y=N],
&
\mathrm{P}
&=
\frac{K}{M},
&
\mathrm{R}
&=
\frac{K}{N},
\nonumber\\
\mathrm{F}_{1}
&=
\frac{2\mathrm{P}\mathrm{R}}
{\mathrm{P}+\mathrm{R}},
&
\mathrm{VOR}
&=
\frac{M}{N}
\label{eq:main_metrics}
\end{align}

Here \(\mathrm{Acc}\) is final-answer exact match, \(\mathrm{P}\) is
precision, \(\mathrm{R}\) is recall, and \(\mathrm{F}_{1}\) is their
harmonic mean. Precision measures supported reported events and
recall measures recovered true events. The Visual Observation Ratio
(VOR) measures count bias. Over eligible positive-event videos \(\mathcal{E}\),
the Accidental Correctness Rate (ACR) and Reasoning Failure Rate
(RFR) are

\begin{align}
\mathrm{ACR}
&=
\frac{1}{|\mathcal{E}|}
\sum_{v\in\mathcal{E}}
\mathbb{I}[\hat y_v=N_v]\,
\mathbb{I}[\mathrm{F}_{1,v}<0.80],
\nonumber\\
\mathrm{RFR}
&=
\frac{1}{|\mathcal{E}|}
\sum_{v\in\mathcal{E}}
\mathbb{I}[\hat y_v\neq N_v]\,
\mathbb{I}[\mathrm{F}_{1,v}\geq0.80]
\label{eq:joint_trace_metrics}
\end{align}

Here \(\mathcal{E}\) is the set of eligible positive-event videos,
\(v\) indexes one video, \(N_v\) is its true event count, \(\hat y_v\)
is its reported final count, and \(\mathrm{F}_{1,v}\) is the trace
score for that video. Thus ACR identifies correct counts unsupported
by a sufficiently faithful trace, whereas RFR identifies incorrect
counts despite a sufficiently faithful trace.

When \(M=0\), precision and \(\mathrm{F}_{1}\) are zero. We average
trace scores over eligible videos; \(\mathrm{VOR}<1\) indicates under-reporting and
\(\mathrm{VOR}>1\) over-reporting.

\subsection{Experimental Protocol and Operating Boundaries}
\label{subsec:experimental_setup}
\begin{figure*}[t]
    \centering
    \begin{subfigure}{\textwidth}
        \centering
        \includegraphics[width=0.9\textwidth]{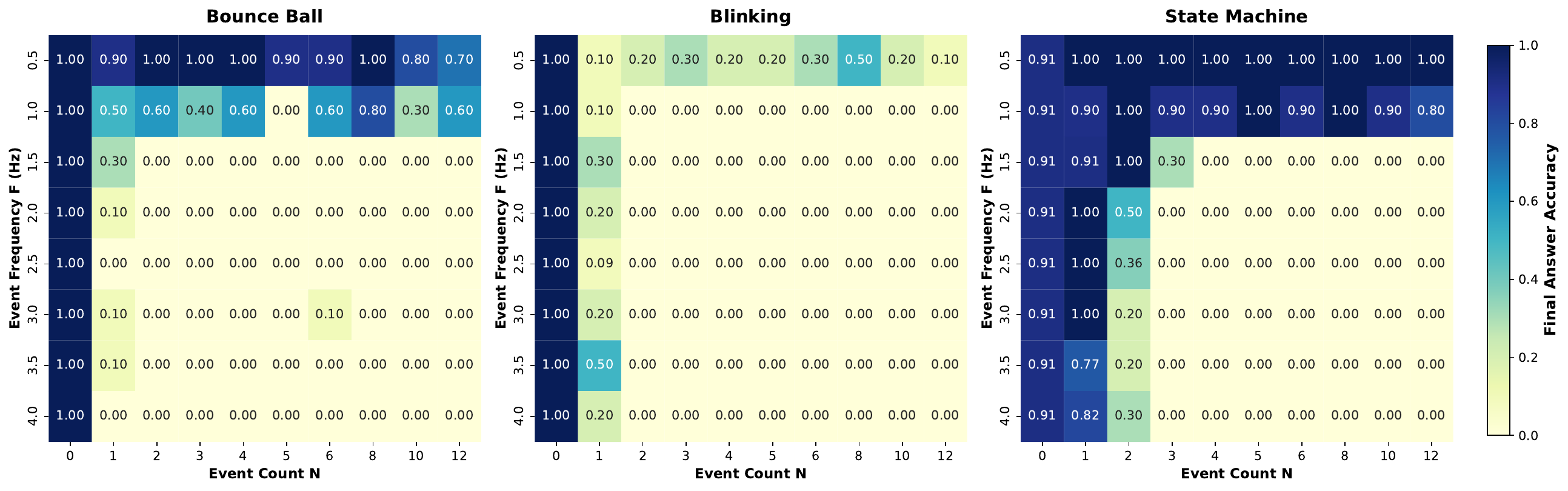}
        \caption{Gemini 3.6 Flash capability surfaces across synthetic domains.}
        \label{fig:main_n_f_surface_gemini}
    \end{subfigure}
    \vspace{0.4em}
    \begin{subfigure}{\textwidth}
        \centering
        \includegraphics[width=0.9\textwidth]{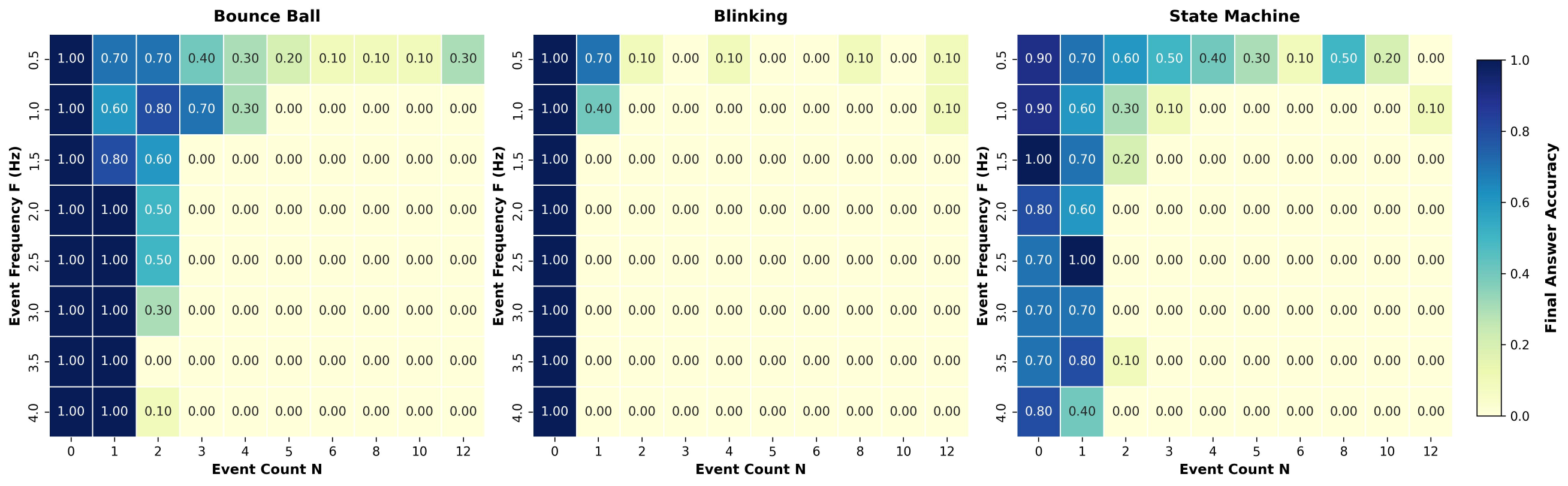}
        \caption{Supplementary Qwen3-VL-235B system profile across synthetic domains.}
        \label{fig:main_n_f_surface_qwen}
    \end{subfigure}
    \caption{\textbf{Capability surfaces across synthetic domains.} Final-answer exact match over event count \(N\) and frequency \(F\) for Bounce Ball, Blinking, and State Machine. Gemini supports the broadest region for persistent State Machine transitions. Qwen shows the same concentration of accuracy at low counts and frequencies.}
    \label{fig:main_n_f_surface}
\end{figure*}

Unparseable final answers are incorrect and malformed traces receive
no matches. Interventions use the same videos and seeds. Sampling
and prompting alter one requested component; event-centered keyframes are an
upper-bound visual-access condition because they may reveal \(N\).

We use \(\tau=0.80\) as a stringent operational reliability target.
A cell must be correct on at least 8 of 10 randomized renderings.
The threshold summarizes the full surface rather than providing a
confidence bound; the cell-wise accuracy surfaces remain available
for inspection. Appendix~\ref{app:variation_and_boundary} reports
seed-level variation and boundary robustness. Let
\(\mathcal{N}_{+}=\mathcal{N}\setminus\{0\}\)
denote positive counts. Here \(d\) indexes a visual domain,
\(\theta\) identifies the evaluated model, and
\(\widehat{\mathcal{P}}_{\theta}^{d}(N,F)\) is its observed exact-match
accuracy from Equation~\ref{eq:parametric_profiling} at count \(N\)
and frequency \(F\). For domain \(d\) and fixed frequency \(F\), the
count boundary is

\begin{equation}
N_{\tau}^{*,d}(F)
=
\max
\left\{
n\in\mathcal{N}_{+}
\ \middle|\
\min_{\substack{n'\in\mathcal{N}_{+}\\n'\leq n}}
\widehat{\mathcal{P}}_{\theta}^{d}(n',F)
\geq\tau
\right\}
\label{eq:operational_count_boundary}
\end{equation}

Here \(n\) is a candidate positive count and \(n'\) ranges over each
tested positive count up to \(n\). The outer maximum selects the
largest candidate whose lowest observed accuracy across that prefix
meets \(\tau\). Thus, it is the largest tested count for which
accuracy remains at least \(80\%\) at that count and every smaller
positive count. If accuracy is below \(80\%\) at \(N=1\), we report
the count boundary as \(N_{\tau}^{*,d}(F)<1\). The \(N=0\) condition
is treated separately as a no-event control.

For domain \(d\) and fixed count \(N\), the complementary
operational frequency boundary is

\begin{equation}
F_{\tau}^{*,d}(N)
=
\max
\left\{
f\in\mathcal{F}
\ \middle|\
\min_{\substack{f'\in\mathcal{F}\\f'\leq f}}
\widehat{\mathcal{P}}_{\theta}^{d}(N,f')
\geq\tau
\right\}
\label{eq:operational_frequency_boundary}
\end{equation}

Here \(f\) is a candidate tested frequency and \(f'\) ranges over
every tested frequency no greater than \(f\). The outer maximum
selects the highest candidate whose lowest observed accuracy across
that lower-frequency prefix meets \(\tau\). It is therefore the
largest tested event rate for which accuracy remains at least
\(80\%\) at that rate and every lower tested rate. If the model fails
at the lowest tested rate of \(0.5\) Hz, we report the frequency
boundary as \(F_{\tau}^{*,d}(N)<0.5\) Hz. Together, the count and
frequency boundaries describe complementary slices of the
two-dimensional reliable operating region.

\section{Mapping Reliable Operating Regions}
\label{sec:capability_boundary_results}
\begin{table*}[t]
\centering
\small
\begin{tabularx}{\textwidth}{X c c c c c c c}
\toprule
\textbf{Evaluation Domain / Region} & \textbf{Final EM (\%)} & \textbf{VOR} & \textbf{Trace P (\%)} & \textbf{Trace R (\%)} & \textbf{Trace F$_1$ (\%)} & \textbf{ACR (\%)} & \textbf{RFR (\%)} \\
\midrule
\multicolumn{8}{l}{\textbf{Part A. Trace Evaluation by Task Domain}} \\
\midrule
  Bounce Ball              &  19.6\% & 0.71 &  59.2\% &  31.5\% &  36.6\% &  3.9\% &  3.6\% \\
  Blinking                 &   6.4\% & 0.60 &  42.2\% &  13.8\% &  18.6\% &  2.4\% &  0.3\% \\
  State Machine            &  37.1\% & 0.74 &  72.5\% &  48.1\% &  53.9\% & 10.8\% &  3.1\% \\
  \midrule
  \textbf{Macro Average}   & \textbf{21.1\%} & \textbf{0.68} & \textbf{58.0\%} & \textbf{31.2\%} & \textbf{36.4\%} & \textbf{5.7\%} & \textbf{2.3\%} \\
\midrule
\multicolumn{8}{l}{\textbf{Part B. Trace Evaluation across ($N \times F$) Capability Regions}} \\
\midrule
  Low Count, Low Freq      &  47.7\% & 1.19 &  50.2\% &  45.6\% &  45.5\% & 15.3\% &  4.2\% \\
  High Count, Low Freq     &  29.2\% & 0.57 &  72.7\% &  46.1\% &  51.8\% &  1.8\% &  4.5\% \\
  Low Count, High Freq     &  17.2\% & 1.12 &  21.7\% &  12.4\% &  14.9\% & 14.2\% &  0.3\% \\
  High Count, High Freq    &   0.2\% & 0.28 &  70.7\% &  18.1\% &  27.7\% &  0.2\% &  0.0\% \\
\midrule
\multicolumn{8}{l}{\textbf{Part C. Trace Diagnosis of Bounce Ball Interventions}} \\
\midrule
\multicolumn{8}{l}{\textit{Visual Evidence Family}} \\
\midrule
  Native + Structured (Baseline)   &  19.6\% & 0.71 &  59.2\% &  31.5\% &  36.6\% &  3.9\% &  3.6\% \\
  Dense Sampling (4 FPS)           &  29.3\% & 0.72 &   4.7\% &   3.3\% &   3.7\% & 28.3\% &  0.4\% \\
  Event-Centered Keyframes         &  68.6\% & 1.07 &  10.0\% &   9.8\% &   9.7\% & 68.3\% &  0.0\% \\
\midrule
\multicolumn{8}{l}{\textit{Prompting and Reasoning Family}} \\
\midrule
  Direct Answer$^\dagger$          &  20.4\% & 1.16 &  45.6\% &  40.2\% &  38.5\% &  7.5\% &  3.2\% \\
  Structured Trace                 &  19.6\% & 0.71 &  59.2\% &  31.5\% &  36.6\% &  3.9\% &  3.6\% \\
  Multi-Turn Verification          &  20.3\% & 0.40 &  15.9\% &   9.5\% &  10.1\% & 18.2\% &  0.6\% \\
  Thinking / CoT                   &  19.5\% & 2.13 &  33.8\% &  48.5\% &  36.9\% & 14.3\% &  2.1\% \\
  Role Prompting                   &  19.3\% & 1.52 &  37.2\% &  44.9\% &  37.6\% & 10.8\% &  2.1\% \\
\bottomrule
\end{tabularx}
\caption{\textbf{Timestamp-recovery evaluation for Gemini 3.6 Flash.} Trace metrics use $\delta(F)=1/(2F)$ and positive-event trials. Final EM includes no-event controls. Part B uses $N\leq3$, $N\geq5$, $F\leq2.0$, and $F\geq2.5$ Hz. Part C evaluates Bounce Ball. $^\dagger$Direct Answer scores voluntarily reported timestamps.}
\label{tab:trace_grounded_evaluation}
\end{table*}

Figure~\ref{fig:main_n_f_surface} reports capability surfaces across
the complete \(N\times F\) grid. Its Gemini panel reports Gemini 3.6 Flash
accuracy. We apply the
Section~\ref{subsec:experimental_setup} boundary definitions with
the \(80\%\) operational target. A reliable cell has at least eight
correct answers from ten randomized renderings. The \(N=0\) column
shows correct no-event responses.

\paragraph{Role of input sampling.}
The profiles evaluate complete end-to-end systems. Gemini receives
\(1\) FPS input and Qwen \(2\) FPS input, so sampling and interface
remain confounds. Higher-density frame sequences in
Section~\ref{sec:diagnostic_interventions} probe whether sampling
contributes to high-frequency failures.

\paragraph{Gemini count boundaries.}
Bounce Ball reaches \(N=10\) only at \(0.5\) Hz and fails the
target at the first positive count from \(1.0\) to \(4.0\) Hz.
Blinking has no reliable positive-count region. State Machine is
reliable through \(N=12\) at \(0.5\) and \(1.0\) Hz, contracts to
\(N=2\) at \(1.5\) Hz, and is generally limited to \(N=1\) above it.

\paragraph{Gemini frequency boundaries.}
Bounce Ball is generally reliable only through \(0.5\) Hz, while
Blinking has no reliable positive-count frequency region. State
Machine reaches \(3.0\) Hz at \(N=1\), \(1.5\) Hz at \(N=2\), and
\(1.0\) Hz from \(N=3\) through \(N=12\). Bounce Ball has an
isolated \(N=8\) exception at \(1.0\) Hz, but its \(N=12\) accuracy
is already below target at the lowest tested frequency. The
contiguous-boundary rule prevents isolated successes after a failure
from being treated as a reliable operating region.

\paragraph{Operating-region interpretation.}
There is no universal count or frequency threshold. Persistent State
Machine transitions yield the broadest region, but the domains also
vary visual form and semantics. With fixed 24-second clips, active
span grows as \(N/F\), so the surfaces profile joint event load and
retention span. They characterize the full model and input pipeline,
not the language model component in isolation.

\paragraph{Supplementary Qwen comparison.}
Figure~\ref{fig:main_n_f_surface_qwen} supplies a supplementary
Qwen3-VL-235B system check. Bounce Ball and Blinking have no broad
reliable region, and State Machine reaches \(N=1\) only in isolated
\(2.5\)- and \(3.5\)-Hz cells. Qwen has no contiguous frequency
region for positive counts because it fails at \(0.5\) Hz. Although
its geometry differs from Gemini, neither system retains a broad
high-count region. Full-resolution capability heatmaps appear in
Appendix~\ref{app:full_heatmaps}, and Appendix~\ref{app:cross_model_results}
reports the model configurations and boundary summaries.

\section{Diagnosing Trace-Level Failures}
\label{sec:trace_grounded_diagnosis}

\begin{figure*}[t]
    \centering
    \includegraphics[width=0.90\textwidth]{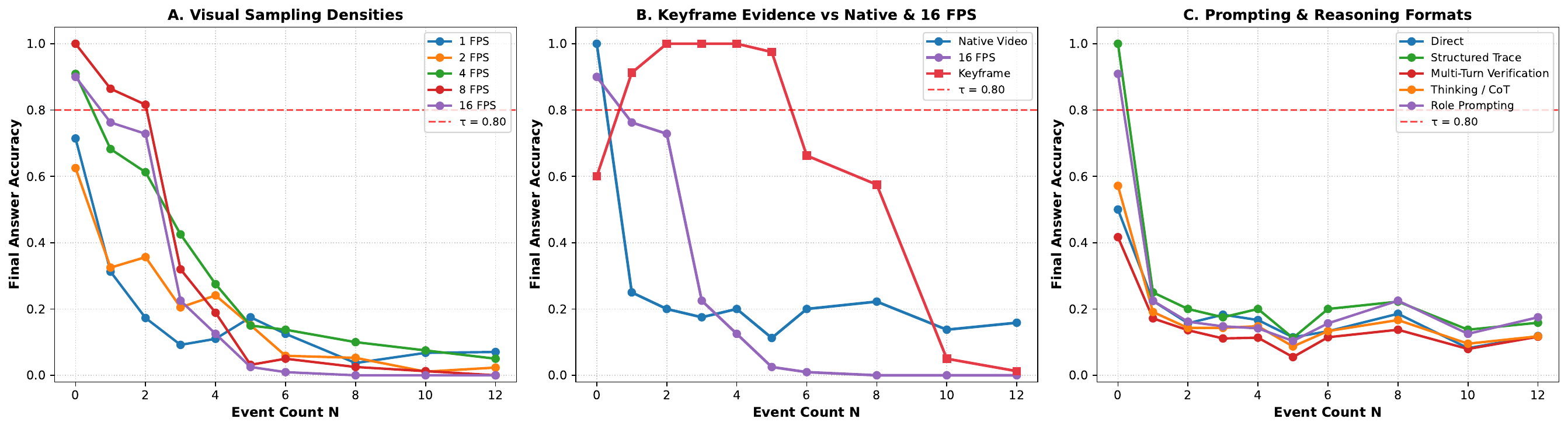}
    \caption{\textbf{Bounce Ball interventions.} Final-answer exact match averaged over frequencies. Sampling helps low counts, event-centered frames help through \(N=5\), and prompting does not expand the reliable region.}
    \label{fig:interventions}
\end{figure*}

Final-answer exact match indicates whether the reported count is
correct, but not whether the model reports timestamps aligned with
the event sequence. Table~\ref{tab:trace_grounded_evaluation}
compares Gemini's reported timestamps with executable traces using
the rate-relative alignment in Section~\ref{subsec:trace_evaluation}.
Part A reports results by visual domain, Part B groups the
\(N\times F\) surface into four operating regions, and Part C
reports the interventions discussed in
Section~\ref{sec:diagnostic_interventions}. These are behavioral
output measures, not direct observations of latent reasoning.

\paragraph{Trace fidelity follows event representation.}
Across domains, macro-average final exact match is \(21.1\%\),
while timestamp recall is \(31.2\%\) and trace \(F_1\) is
\(36.4\%\). State Machine performs best in both final prediction
and timestamp recovery, with \(37.1\%\) exact match and \(53.9\%\)
trace \(F_1\). Blinking performs worst, with \(6.4\%\) exact match
and \(18.6\%\) trace \(F_1\). Bounce Ball lies between these
extremes. This ordering matches the operating regions in
Figure~\ref{fig:main_n_f_surface_gemini}. It is consistent with
persistent event displays producing more recoverable reported
timestamps, while not isolating persistence from other domain
differences.

\paragraph{The error pattern changes with count and frequency.}
In the low-count, low-frequency region, the Visual Observation
Ratio is \(1.19\), so the model reports more timestamps than occur.
Precision is \(50.2\%\), and \(15.3\%\) of trials have a correct
final answer despite a low-fidelity reported trace. Some correct
counts therefore coexist with unmatched reported timestamps.

At high count and low frequency, the Visual Observation Ratio
falls to \(0.57\), indicating substantial under-reporting.
Precision rises to \(72.7\%\), but recall remains only
\(46.1\%\). The model is usually accurate about the timestamps it
does report, while omitting much of the sequence.

This effect is strongest at high count and high frequency.
The model reports only \(28\%\) as many timestamps as occur, and
trace recall falls to \(18.1\%\), while precision remains
\(70.7\%\). The dominant high-load signature is therefore
incomplete reported-timestamp recovery rather than broad
over-reporting. Separating missed events from merged events would
require a richer type- and order-aware alignment than used here.

The regional comparison reveals two distinct output regimes. At low
load, excess reports and accidental correct counts are consistent
with a mixture of unmatched and compensating events. At high load,
reported events are comparatively well supported but far too few,
which is consistent with omission or merging dominating the failure.
The final integer alone would obscure this shift in error pattern.

\paragraph{Failure can also occur after event recovery.}
Most degradation is associated with incomplete reported traces, but
timestamp recovery does not fully determine the final answer. RFR
reaches \(4.5\%\) in the high-count, low-frequency region. In these
cases, the reported trace meets the \(F_1\) criterion but the final
count remains incorrect, which is consistent with an output-level
trace-to-answer aggregation mismatch.

\section{What Moves the Boundary?}
\label{sec:diagnostic_interventions}

We choose Bounce Ball for interventions because each wall contact
requires spatial localization of an interaction as well as temporal
separation, retention, and accumulation of repeated events. We test
whether its failures arise from temporal sampling, event localization,
or inference format. Figure~\ref{fig:interventions} reports accuracy
averaged across frequencies. Appendix~\ref{app:visual_access_interventions} contains
the complete intervention surfaces, and
Table~\ref{tab:trace_grounded_evaluation} Part C contains trace metrics.

\paragraph{Denser sampling improves low-count temporal access.}
Gemini's native-video interface uses approximately \(1\) FPS in
our setup. Explicit \(4\)-FPS sampling raises overall accuracy from
\(19.6\%\) to \(29.3\%\), and \(10\) FPS is reliable for one event
through \(4.0\) Hz. The recovery does not extend to longer sequences,
and no stable high-count region emerges. The \(4\)-FPS trace
\(F_1\) is only \(3.7\%\), while \(28.3\%\) of trials have a correct
answer with a low-fidelity trace. Better counts therefore need not
reflect faithful event recovery.

\paragraph{Keyframe evidence produces the largest recovery.}
For each ground-truth event at \(t_i\), we provide keyframes at
\(t_i-0.1\), \(t_i\), and \(t_i+0.1\) seconds. This removes event
search from the complete video. Accuracy rises to \(68.6\%\) and is
near-perfect through \(N=5\), but falls to \(0.66\) at \(N=6\),
\(0.57\) at \(N=8\), and nearly zero at \(N\geq10\). Localization is
therefore important, but retention or accumulation still limits
longer sequences. The trace \(F_1\) remains \(9.7\%\), and \(68.3\%\)
of trials have a correct answer with a low-fidelity trace. Because
one frame group is supplied per true event, this is an upper-bound
visual-access condition.

\paragraph{Reasoning formats do not provide consistent recovery.}
Direct answering, structured tracing, multi-turn verification,
thinking, and role prompting yield \(19.3\%\) to \(20.4\%\) accuracy.
Thinking raises recall to \(48.5\%\) but over-reports events with a
Visual Observation Ratio of \(2.13\) and \(33.8\%\) precision.
Structured tracing is more precise at \(59.2\%\), but recovers only
\(31.5\%\) of true events. No prompting format expands the reliable
count or frequency region.

Together, the interventions show that improved visual access does not
eliminate the high-count boundary.

\section{Natural-Video Boundary Transfer}
\label{sec:real_world_transfer}

\begin{figure}[h]
    \centering
    \includegraphics[width=\linewidth]{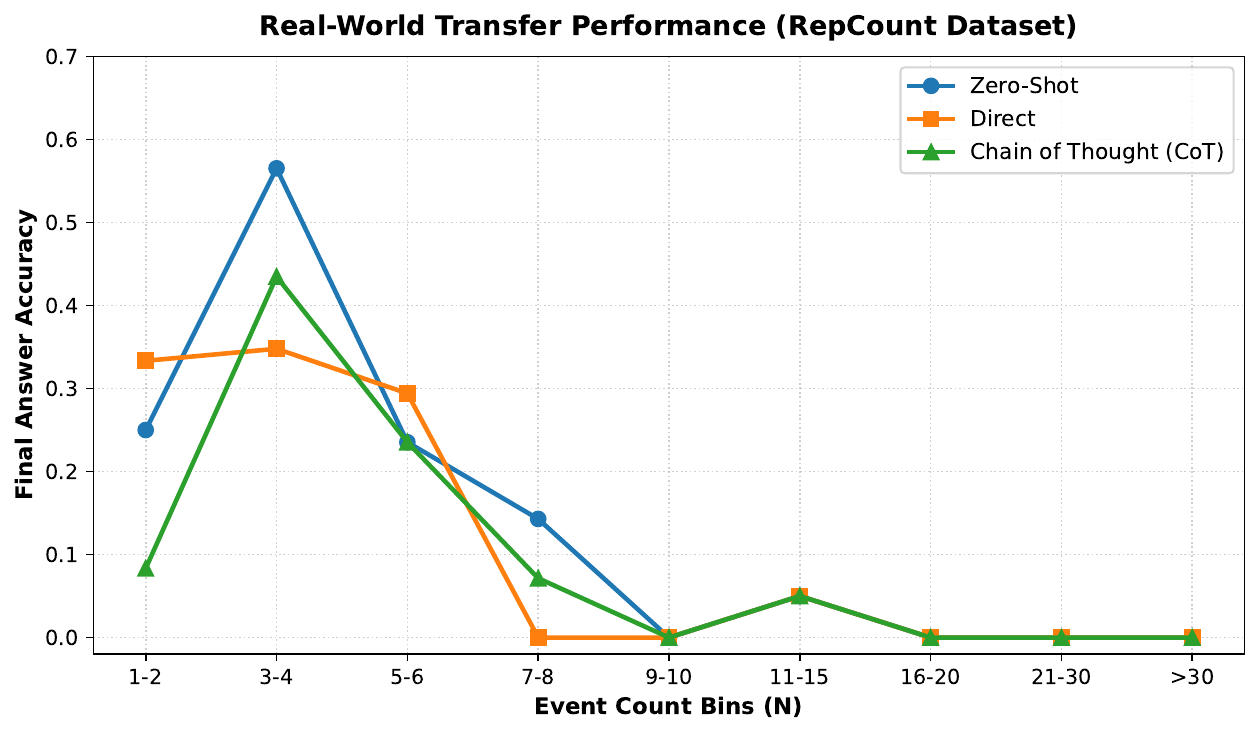}
    \caption{\textbf{Natural-video count-pattern check.} Exact-match accuracy on 152 TransRAC clips \citep{hu2022transrac}. All prompting formats concentrate accuracy in low-count bins.}
    \label{fig:real_world_transfer}
\end{figure}

Figure~\ref{fig:real_world_transfer} reports exact-match accuracy by
event-count bin for the three prompting formats. We use 151 clips from TransRAC with repetition annotations
\citep{hu2022transrac}, grouped by event count. Each clip is
evaluated once with zero-shot, direct-answer, and chain-of-thought
prompting for 456 trials. These natural videos contain camera motion,
clutter, occlusion, and variable lighting. Frequency cannot be
controlled independently, so this is a count-pattern check rather
than a replication of the \(N\times F\) surface. It reports
final-answer exact match by count bin and does not apply trace-level
evaluation. Accuracy concentrates in low-count bins \(N\leq4\), and
no prompting format establishes a stable high-count region. This
supports the qualitative prediction that reliability declines as
repeated-event load rises despite natural variation, without claiming
matched absolute accuracy or a specific causal mechanism. Natural
videos therefore test external relevance of the controlled pattern,
not whether the synthetic and natural distributions are identical.
Outcomes above six events are sparse and often zero across prompting
formats.

\section{Conclusion and Discussion}
\label{sec:discussion_and_conclusion}

\paragraph{Beyond benchmark scoring.}
This work maps reliable temporal regions and intervention
effects. Controlled \(N\times F\) profiles separate visibility,
resolution, sequence length, and aggregation.

\paragraph{What the diagnosis reveals.}
Results are consistent with staged failure involving event access,
retention, and aggregation. State Machine is easier than Blinking,
but performance declines with sequence length. Keyframes provide one
frame group per event, yet their residual decline shows visual search
is insufficient.

\paragraph{Why executable traces and interventions matter.}
Final answers cannot distinguish access from downstream failures.
Alignment reveals unmatched, under-reported, and over-reported events;
joint metrics expose answer-trace mismatches. Interventions test
alternatives.

\paragraph{Limitations.}
The study uses clean, regularly spaced repeated events, and increasing
\(N\) at fixed \(F\) raises event load and span. Only two systems are
profiled synthetically. Natural-video evaluation neither controls
frequency nor uses executable traces. Reported traces are behavioural
evidence, not direct access to latent computation.

\section*{Acknowledgments}
Baskar, Cai, Shabihi, Satheesh, Islam, and Huang are supported by DARPA HR001124S0029-AIQ-FP-019, National Science Foundation TRAILS Institute (2229885). Private support was provided by Open Philanthropy and Apple. The Authors acknowledge the National Artificial Intelligence Research Resource (NAIRR) Pilot for contributing to this research result.

\bibliography{aaai2027}

\clearpage
\appendix
\makeatletter
\renewcommand{\addcontentsline}[3]{%
  \addtocontents{#1}{\protect\contentsline{#2}{#3}{\thepage}{}}%
}
\makeatother
\setcounter{tocdepth}{2}
\tableofcontents
\section*{Appendix Overview}
\noindent
This appendix provides the extended evidence, methodological detail, and
reproducibility material that support the main paper's central finding: current
video--language models can fail at elementary event bookkeeping as temporal
load increases. The sections below connect each supplement directly to a
component of the main-paper argument.

\begin{itemize}
    \item \textbf{Section~A} specifies the controlled Manim-based task renderer,
    supplied frame sequences, and model configurations. These details establish
    that the capability boundary is measured on videos with known temporal ground truth.
    \item \textbf{Section~B} provides the full capability heatmaps, cross-model
    boundary summaries, seed-level variation, and operational boundary robustness,
    showing how performance changes across event count, frequency, scale, and domain.
    \item \textbf{Section~C} details diagnostic interventions (prompting formats,
    sampling density sweeps, and oracle keyframes) alongside trace extraction,
    matching algorithms, and error auditing.
    \item \textbf{Section~D} provides the complete computational API financial cost analysis
    across all experiments and visual domains.
    \item \textbf{Section~E} situates the benchmark within the broader video-language
    and programmatic video literature.
    \item \textbf{Section~F} provides representative qualitative trace examples across all 6 diagnostic failure categories.
\end{itemize}

\section{Benchmark Setup and Model Specifications}
\label{app:synthetic_generation_details}

\subsection{Task and Renderer Specifications}
\label{app_sub:task_renderer_specifications}

Each synthetic instance is specified by a visual domain \(d\), event
count \(N\), event frequency \(F\), and rendering seed \(s\). We use
Bounce Ball wall contacts, Blinking on-pulses, and State Machine
categorical transitions. The task question asks for the number of
target events in the rendered video.

\paragraph{Manim Animation Engine and Trace Generation.}
We construct all synthetic video benchmarks using the Manim\footnote{\url{https://github.com/3b1b/manim}} mathematical animation engine in Python. Manim is chosen because it allows programmatic, frame-accurate control over spatial trajectories, visual state transitions, and rendering framerates without visual compression artifacts or timing jitter. For each trial \((d, N, F, s)\):
\begin{enumerate}
    \item \textbf{Event Scheduling:} Given count $N$ and frequency $F$, a deterministic event scheduler calculates the ground-truth event timestamps $t_1, t_2, \dots, t_N$ within the active temporal window.
    \item \textbf{Manim Video Rendering:} Manim scenes procedurally animate visual elements (wall collisions in \texttt{bounce\_ball}, visual pulse opacities in \texttt{blinking}, and categorical node highlights in \texttt{state\_machine}) matching the exact timestamps $t_i$.
    \item \textbf{Executable Trace Generation:} Simultaneously during scene construction, the renderer logs an executable ground-truth trace
    \[
    \mathcal{T}^{*}=\bigl((t_i,e_i,s_i^{-},s_i^{+},c_i)\bigr)_{i=1}^{N},
    \]
    where \(t_i\) is the event timestamp, \(e_i\) is the event type, \(s_i^{-}\) and \(s_i^{+}\) are the states before and after the event, and \(c_i\) is the running cumulative count. Executing the trace returns the final count \(c_N=N\).
\end{enumerate}

We evaluate
\(
N\in\{0,1,2,3,4,5,6,8,10,12\}
\)
and
\(
F\in\{0.5,1.0,1.5,2.0,2.5,3.0,3.5,4.0\}\,\mathrm{Hz}.
\)
Every clip has a fixed duration of 24 seconds. The active event
sequence occupies approximately \(N/F\) seconds. The remaining time
is filled with static frames before or after the active sequence, with
the allocation determined by the rendering seed. Thus the design
varies event load and active temporal span without changing total
duration.

For each positive-count \((N,F)\) cell, we render ten randomized
videos. The seed changes nuisance appearance factors, including color,
geometry, trajectory, initial state, orientation, and start delay,
while preserving the event schedule. The \(N=0\) control is rendered
ten times at nominal \(F=1.0\,\mathrm{Hz}\) and is excluded from
positive-event trace ratios. This yields 730 videos per domain and
2,190 videos across the three domains.

\paragraph{Rationale for Capability Axes, Parameter Ranges, and Granularity.}
We select Event Count ($N$) and Event Frequency ($F$) as orthogonal capability axes to isolate two complementary failure modes in video-language models: temporal working memory load ($N$) and temporal perceptual resolution ($F$). The frequency range $F \in [0.5, 4.0]\text{ Hz}$ spans from $0.5\text{ Hz}$ (consecutive events separated by $2.0\text{s}$, well within standard VLM 1--2 FPS sampling windows) up to $4.0\text{ Hz}$ ($0.25\text{s}$ spacing), deliberately pushing models beyond standard VLM frame-sampling resolution. The count range $N \in [0, 12]$ includes $N=0$ as a no-event negative control and extends up to $N=12$, which fills the entire 24-second clip duration at the minimum frequency $F=0.5\text{ Hz}$ ($N/F = 12 / 0.5 = 24\text{s}$). Sweep granularity uses finer steps at low event counts ($N \in \{1, 2, 3, 4, 5, 6\}$) to precisely locate the onset of breakdown, and wider steps at high loads ($N \in \{8, 10, 12\}$). This parametric design generalizes to any temporal accounting task by defining a target event primitive, sweeping event load $N$ and occurrence rate $F$, and auditing predicted responses against executable ground-truth traces.

\subsection{Supplied Frame Sequences}
\label{app_sub:supplied_frame_sequences}

The reported sampling rates describe the frames supplied to the
evaluated systems, not undocumented internal video processing.
Gemini receives native video at approximately 1 FPS in our setup and
Qwen receives 2 FPS. The Bounce Ball visual-access study reuses the
same generated videos and seed identifiers while changing one source
of evidence at a time. Frame-sampling conditions provide regularly
sampled frame sequences at the selected density. The event-centered
keyframe condition supplies three frames around each true event,
at \(t_i-0.1\), \(t_i\), and \(t_i+0.1\) seconds. This condition
removes the need to search the full video for event locations, but it
still requires the model to retain and aggregate the supplied event
evidence. The frame interventions therefore diagnose whether a change
in final-answer accuracy is compatible with improved external visual
access, without attributing the change to any particular internal
mechanism.

\subsection{Model Configurations}
\label{app_sub:model_configurations}

The main synthetic evaluation compares Gemini 3.6 Flash and
Qwen3-VL-235B. Both systems receive the same domain-specific zero-shot
question and use their native video interfaces. Gemini is supplied
approximately 1 FPS and Qwen 2 FPS. These rates are external input
settings. They should not be interpreted as measurements of either
system's internal temporal processing.

Table~\ref{tab:app_cross_model_summary} reports the configuration,
domain-level exact-match accuracy, and maximum count boundary for the
available cross-model profiles. The table shows that State Machine is
the strongest domain for every listed model, while the reliable count
prefix remains short for most open-model configurations. The 38B
InternVL variant reaches the largest open-model count boundary of six,
but this remains below Gemini's boundary of twelve.

\begin{table*}[t]
\centering
\small
\begin{tabular}{lccccc}
\toprule
\textbf{Model} & \textbf{Input} & \textbf{Bounce EM (\%)} & \textbf{Blink EM (\%)} & \textbf{State EM (\%)} & \textbf{Max. count boundary} \\
\midrule
Gemini 3.6 Flash & 1 FPS & 19.6 & 6.4 & 37.1 & 12 \\
\midrule
Qwen3-VL-8B & 2 FPS & 12.4 & 10.9 & 26.6 & 2 \\
Qwen3-VL-32B & 2 FPS & 15.5 & 13.2 & 26.2 & 1 \\
Qwen3-VL-235B & 2 FPS & 19.2 & 12.0 & 26.4 & 1 \\
\midrule
InternVL3.5-8B & 16 frames & 9.0 & 11.2 & 26.2 & 3 \\
InternVL3.5-30B-A3B & 16 frames & 9.2 & 10.4 & 24.5 & 3 \\
InternVL3.5-38B & 16 frames & 12.8 & 9.9 & 32.0 & 6 \\
\bottomrule
\end{tabular}
\caption{\textbf{Cross-model temporal-profile summary.} Exact-match accuracy is reported separately for Bounce Ball, Blinking, and State Machine across 730 trials per domain. The maximum count boundary is the largest reliable contiguous positive-count prefix observed at any tested frequency and domain.}
\label{tab:app_cross_model_summary}
\end{table*}

Table~\ref{tab:app_cross_model_trace} compares the main-paper Gemini
macro average with pooled open-model trace metrics. Although the two
render sets differ slightly, both instantiate the same three tasks and
the same \(N\)-by-\(F\) design. The comparison is therefore
task-matched rather than paired. As detailed in
Appendix~\ref{app:trace_evaluation_and_validation}, these statistics
evaluate timestamps expressed in the model response and constitute
behavioural evidence rather than direct measurements of internal
representations.

\begin{table*}[t]
\centering
\small
\begin{tabular}{lrrrrrrr}
\toprule
\textbf{Model} & \textbf{EM (\%)} & \textbf{VOR} & \textbf{P (\%)} & \textbf{R (\%)} & \textbf{Trace \(F_1\) (\%)} & \textbf{ACR (\%)} & \textbf{RFR (\%)} \\
\midrule
Gemini 3.6 Flash & 21.1 & 0.68 & 58.0 & 31.2 & 36.4 & 5.7 & 2.3 \\
\midrule
Qwen3-VL-8B & 16.6 & 13.04 & 21.0 & 18.8 & 15.8 & 6.9 & 1.4 \\
Qwen3-VL-32B & 18.3 & 0.43 & 21.2 & 14.0 & 14.8 & 9.3 & 1.5 \\
Qwen3-VL-235B & 19.2 & 0.21 & 18.4 & 10.0 & 11.4 & 10.3 & 1.1 \\
\midrule
InternVL3.5-8B & 15.5 & 0.00 & 0.0 & 0.0 & 0.0 & 15.5 & 0.0 \\
InternVL3.5-30B-A3B & 14.7 & 0.00 & 0.0 & 0.0 & 0.0 & 13.7 & 0.0 \\
InternVL3.5-38B & 18.2 & 0.00 & 0.0 & 0.0 & 0.0 & 13.8 & 0.0 \\
\bottomrule
\end{tabular}%
\caption{\textbf{Task-matched cross-model trace metrics across temporal domains.} Macro averages reported on task-matched videos, with 730 trials per domain. Metrics score only explicit, parser-recognized timestamps and do not infer events from free-form text.}
\label{tab:app_cross_model_trace}
\end{table*}

Gemini attains the highest pooled trace \(F_1\) (\(36.4\%\)). Among
the open models, Qwen3-VL-8B attains the highest reported-trace
\(F_1\) (\(15.8\%\)), but exhibits severe over-reporting
(VOR \(=13.04\)). Qwen3-VL-32B and Qwen3-VL-235B instead
under-report events (VOR \(=0.43\) and \(0.21\), respectively).
InternVL yields
nonzero final-answer accuracy but no extractable timestamped events.
Accordingly, its zero trace values characterize the format of its
visible responses, rather than establishing that the model observed no
events. A controlled cross-model trace-fidelity comparison would
require a common structured-output prompt; the present results compare
the evidence each model externalizes under its available protocol.

\section{Capability Boundaries and Statistical Robustness}
\label{app:full_heatmaps}

This section provides full-resolution capability surfaces for the
primary evaluation and extended model-family comparisons. Each cell
reports final-answer exact match for a controlled event-count and
event-frequency condition. The family plots provide a qualitative
cross-architecture check of whether the low-count concentration
observed in the primary profiles recurs at other model scales.
Figures~\ref{fig:app_full_qwen_heatmaps} and
\ref{fig:app_full_internvl_heatmaps} provide the Qwen3-VL family and
InternVL3.5 family profiles, respectively.

\subsection{Qwen3-VL Scale Comparison}

\begin{figure*}[p]
    \centering
    \includegraphics[width=0.94\textwidth]{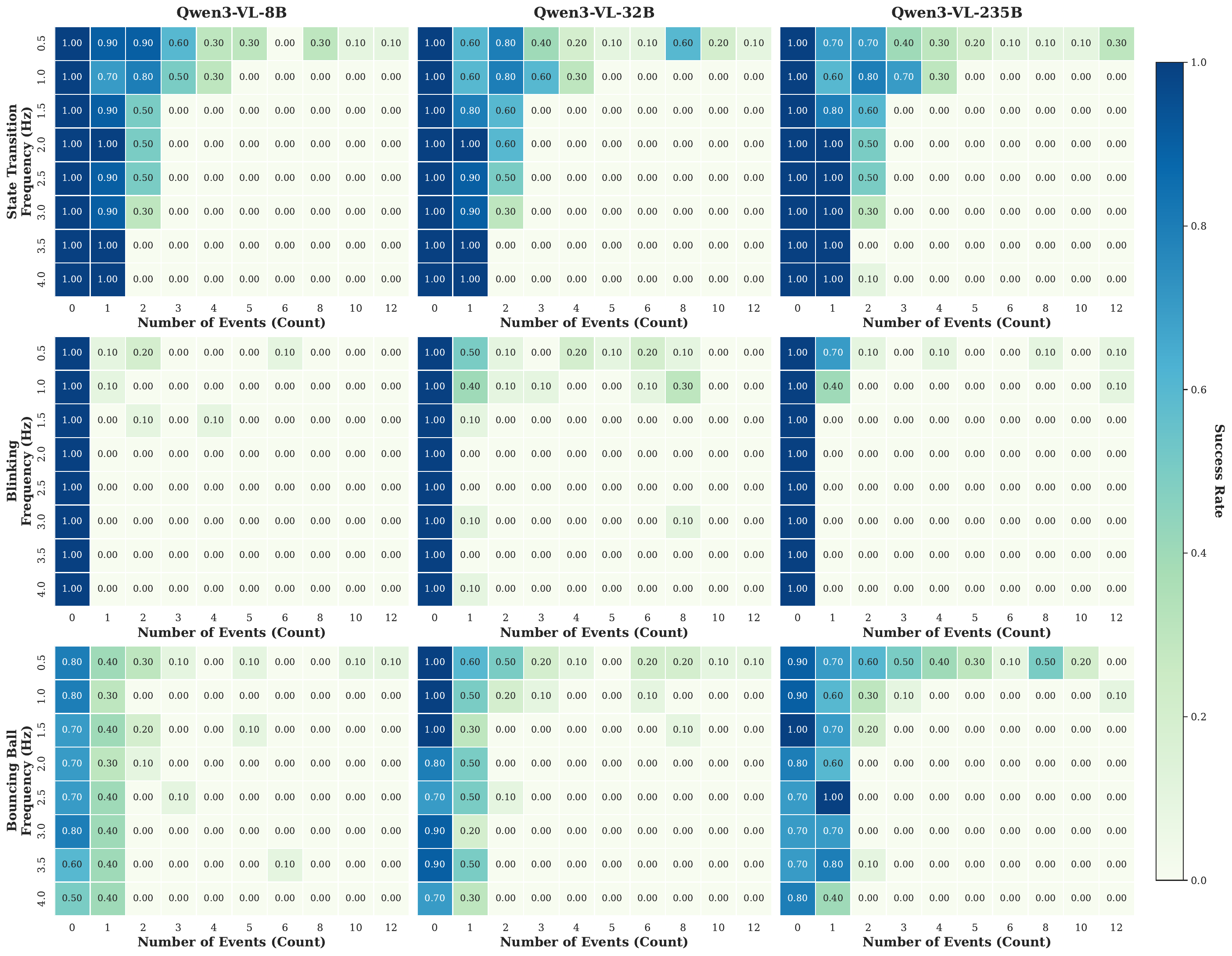}
    \caption{\textbf{Qwen3-VL family capability heatmaps.} Final-answer exact match across event count and frequency for State Machine transitions. The panels compare Qwen3-VL-8B, Qwen3-VL-32B, and Qwen3-VL-235B.}
    \label{fig:app_full_qwen_heatmaps}
\end{figure*}

Figure~\ref{fig:app_full_qwen_heatmaps} compares Qwen3-VL-8B,
Qwen3-VL-32B, and Qwen3-VL-235B on the State Machine task. Across
the three scales, accuracy is concentrated at small event counts and
falls sharply once the count exceeds four, including at the lowest
tested frequency. Larger scales improve some easy cells, but the
figure does not show a broad high-count region at slow event rates.
Thus, scale changes the level of accuracy in parts of the surface but
does not remove the count-dependent boundary in this task.

\subsection{InternVL3.5 Scale and Domain Comparison}

\begin{figure*}[p]
    \centering
    \includegraphics[width=0.94\textwidth]{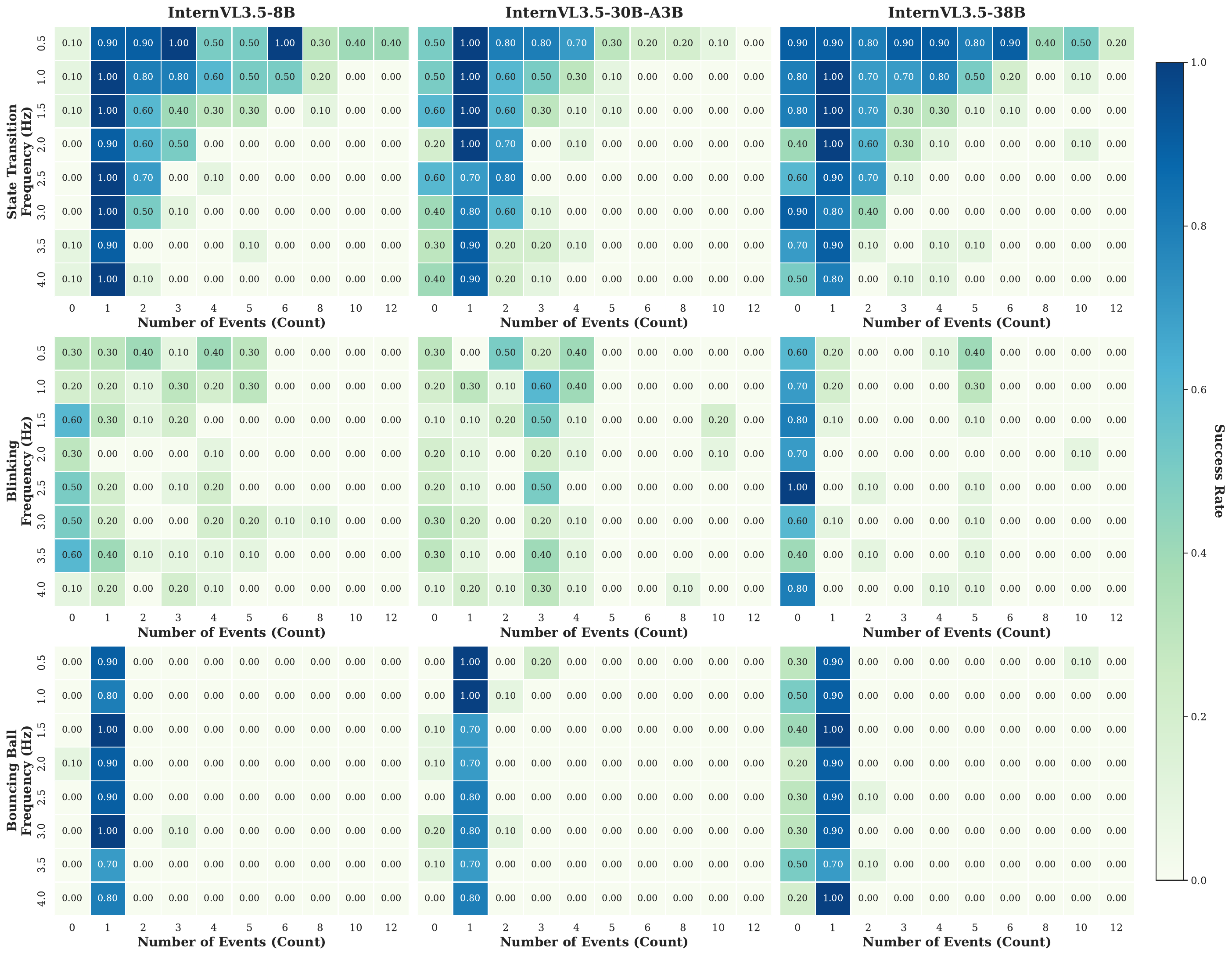}
    \caption{\textbf{InternVL3.5 family capability heatmaps.} Final-answer exact match across event count and frequency for State Machine, Blinking, and Bounce Ball. Columns compare the evaluated InternVL3.5 scales.}
    \label{fig:app_full_internvl_heatmaps}
\end{figure*}

Figure~\ref{fig:app_full_internvl_heatmaps} extends the comparison
to InternVL3.5-8B, InternVL3.5-30B-A3B, and InternVL3.5-38B across
State Machine, Blinking, and Bounce Ball. The family shares the same
qualitative shape across domains. Performance is strongest in the
small-count regime and contracts as event count grows, including in
low-frequency conditions where events are well separated. These
figures report end-to-end system behaviour under controlled inputs;
they do not identify the internal mechanism producing the boundary.
Their value is to show that the boundary pattern is not restricted to
one Qwen configuration or to one temporal domain.

\subsection{Operational Boundary Summary}
\label{app_sub:boundary_summary}
\label{app:cross_model_results}

For each model, domain, and fixed frequency, the count boundary is the
largest positive count whose tested prefix meets the \(80\%\)
operational target. The frequency boundary is defined analogously at a
fixed count. This contiguous-prefix rule prevents a later isolated
success from being reported as a reliable operating region. We use the
cross-model comparison as a qualitative robustness check, not as a
claim of universality across video-language models.

\subsection{Seed-Level Variation}
\label{app_sub:seed_level_variation}

Each synthetic cell contains ten independently randomized renderings.
For a cell with \(k\) correct answers, the reported accuracy is
\(k/10\). Seed-level uncertainty is summarized with a two-sided 95\%
Wilson score interval for selected headline cells and intervention
comparisons. The full heatmaps retain the cell-wise accuracy rather
than replacing the surface with a single aggregate score.

Table~\ref{tab:app_seed_variation} reports the interval summary for
the baseline and key intervention comparisons. The baseline interval
does not overlap the event-centered-keyframe interval, and the
keyframe improvement is much larger than the corresponding sampling
improvement. These intervals quantify uncertainty in the aggregate
exact-match estimates, while the heatmaps retain the variation across
individual \((N,F)\) cells.

\begin{table*}[t]
\centering
\small
\begin{tabular}{lrrrrr}
\toprule
\textbf{Condition} & \textbf{Correct Runs} & \textbf{Total Runs} & \textbf{Accuracy (\%)} & \textbf{Lower Interval (\%)} & \textbf{Upper Interval (\%)} \\
\midrule
Gemini native-video baseline & 143 & 730 & 19.6 & 16.9 & 22.6 \\
Gemini 4-FPS sampling & 214 & 730 & 29.3 & 26.1 & 32.7 \\
Gemini event-centered keyframes & 501 & 730 & 68.6 & 65.2 & 71.9 \\
\bottomrule
\end{tabular}
\caption{\textbf{Gemini seed-level variation for Bounce Ball interventions.} Each condition has 730 trials. Intervals are two-sided 95\% Wilson score intervals for final-answer exact match.}
\label{tab:app_seed_variation}
\end{table*}

\subsection{Operational Boundary Robustness}
\label{app_sub:boundary_robustness}
\label{app:variation_and_boundary}

We use \(\tau=0.80\) as an operational reliability target. A cell is
reliable when at least eight of ten renderings are correct. At fixed
frequency, the count boundary is the largest tested positive count for
which that count and every smaller tested positive count meet the
target. At fixed count, the frequency boundary applies the same rule
over increasing tested frequencies. This definition is descriptive and
does not treat the threshold as a confidence bound.

\section{Interventions, Prompts, and Trace Evaluation}
\label{app:prompting_details}

This appendix documents all domain task questions, prompt condition templates, and input formatting instructions across the 5 core experiments defined in our controlled evaluation framework.

\subsection{Domain Task Questions}
\label{app_sub:domain_questions}

For every experimental trial across all 5 experiments, the model receives the visual video input paired with a zero-shot domain task question requesting the total count of target events. The exact questions per benchmark domain are:

\begin{promptbox}{Benchmark Task Questions}
\textbf{Bounce Ball Domain:}\\
"How many times did the ball contact the walls?"\\[6pt]
\textbf{Blinking Domain:}\\
"How many times did the object blink?"\\[6pt]
\textbf{State Machine Domain:}\\
"How many state transitions occurred in the video?"
\end{promptbox}

\subsection{Experiment 1: Full $N \times F$ Matrix Boundary Sweep}
\label{app_sub:exp1_prompts}

\paragraph{Motivation \& Protocol.} Experiment 1 establishes the baseline operational capability boundary ($x^*$) across Event Count ($N \in [0, 1, 2, 3, 4, 5, 6, 8, 10, 12]$) and Event Frequency ($F \in [0.5, 1.0, 1.5, 2.0, 2.5, 3.0, 3.5, 4.0]\text{ Hz}$). All trials in Experiment 1 use the \texttt{structured\_trace} prompt baseline on native video inputs.

\begin{promptbox}{Experiment 1: Structured Trace Baseline Prompt (\texttt{structured\_trace})}
Question: {question}

Observe the video carefully and provide a step-by-step reasoning trace tracking every event, its timestamp (in seconds), and the running count.
Format your response as follows:
1. Step-by-Step Event Ledger (timestamp, event type, running count)
2. Final Answer inside \textbackslash boxed\{\} (e.g. \textbackslash boxed\{4\})
\end{promptbox}

\subsection{Experiment 2: Frame Sampling Density Interventions}
\label{app_sub:exp2_prompts}

\paragraph{Motivation \& Protocol.} Experiment 2 tests whether performance breakdown at high frequencies ($F \ge 2.0\text{ Hz}$) stems from temporal perceptual sampling limits or downstream temporal reasoning limits. It evaluates 7 sampling densities (\texttt{native\_video}, 1, 2, 4, 8, 10, 16~FPS) paired with the structured trace prompt:

\begin{promptbox}{Experiment 2: Frame Sampling Density Intervention Prompt}
[INPUT MODE: Supplied Frame Grid at {FPS} FPS]

Question: {question}

Observe the video carefully and provide a step-by-step reasoning trace tracking every event, its timestamp (in seconds), and the running count.
Format your response as follows:
1. Step-by-Step Event Ledger (timestamp, event type, running count)
2. Final Answer inside \textbackslash boxed\{\} (e.g. \textbackslash boxed\{4\})
\end{promptbox}

\subsection{Experiment 3: Oracle Keyframe Evidence Interventions}
\label{app_sub:exp3_prompts}

\paragraph{Motivation \& Protocol.} Experiment 3 isolates video frame search from temporal tallying by supplying perfect oracle visual keyframes. For each ground-truth event at $t_i$, frame triplets at $(t_i - 0.1\text{s}, t_i, t_i + 0.1\text{s})$ are extracted and provided directly to the model.

\begin{promptbox}{Experiment 3: Oracle Keyframe Evidence Prompt (\texttt{oracle\_evidence})}
[ORACLE EVIDENCE AVAILABLE: Key event windows extracted at $t_i \pm 0.1\text{s}$]

Question: {question}

Observe the video carefully and provide a step-by-step reasoning trace tracking every event, its timestamp (in seconds), and the running count.
Format your response as follows:
1. Step-by-Step Event Ledger (timestamp, event type, running count)
2. Final Answer inside \textbackslash boxed\{\} (e.g. \textbackslash boxed\{4\})
\end{promptbox}

\subsection{Experiment 4: Prompting \& Reasoning Mode Interventions}
\label{app_sub:exp4_prompts}

\paragraph{Motivation \& Protocol.} Experiment 4 measures the impact of prompt structures, reasoning formats, self-correction, and system instructions on capability breakdown. It evaluates 5 distinct prompt conditions:

\begin{promptbox}{Experiment 4.1: Direct Answer Condition (\texttt{direct})}
Question: {question}

Observe the video carefully. Provide your final answer formatted inside \textbackslash boxed\{\} (e.g. \textbackslash boxed\{4\}).
\end{promptbox}

\begin{promptbox}{Experiment 4.2: Structured Trace Condition (\texttt{structured\_trace})}
Question: {question}

Observe the video carefully and provide a step-by-step reasoning trace tracking every event, its timestamp (in seconds), and the running count.
Format your response as follows:
1. Step-by-Step Event Ledger (timestamp, event type, running count)
2. Final Answer inside \textbackslash boxed\{\} (e.g. \textbackslash boxed\{4\})
\end{promptbox}

\begin{promptbox}{Experiment 4.3: Multi-Turn Verification Condition (\texttt{multi\_turn\_verification})}
Question: {question}

First, carefully list every detected event with its approximate timestamp and current count.
Second, audit your event list to verify whether any events were missed, merged, or double-counted.
Finally, state your verified final answer inside \textbackslash boxed\{\} (e.g. \textbackslash boxed\{4\}).
\end{promptbox}

\begin{promptbox}{Experiment 4.4: Thinking / Chain-of-Thought Condition (\texttt{thinking})}
Question: {question}

Use deep step-by-step video analysis. Thoroughly analyze the temporal video sequence frame-by-frame, count all target visual events, and output your final count inside \textbackslash boxed\{\} (e.g. \textbackslash boxed\{4\}).
\end{promptbox}

\begin{promptbox}{Experiment 4.5: Role Prompting Condition (\texttt{role\_prompting})}
System Instruction: You are an expert video analytics systems auditor specializing in fine-grained temporal event verification.

Question: {question}

Apply rigorous visual event auditing. Log all timestamped visual transitions and state your final answer inside \textbackslash boxed\{\} (e.g. \textbackslash boxed\{4\}).
\end{promptbox}

\subsection{Experiment 5: MORSE Trace Diagnosis \& Error Taxonomy}
\label{app_sub:exp5_prompts}

\paragraph{Motivation \& Protocol.} Experiment 5 audits intermediate model reasoning traces against executable ground-truth traces to measure Trace Precision ($P$), Recall ($R$), Trace $F_1$, Accidental Correctness Rate (ACR), and Reasoning Failure Rate (RFR).

\begin{promptbox}{Experiment 5: MORSE Executable Trace Diagnostic Prompt}
Question: {question}

Observe the video carefully and provide a step-by-step reasoning trace tracking every event, its timestamp (in seconds), and the running count.
Format your response as follows:
1. Step-by-Step Event Ledger (timestamp, event type, running count)
2. Final Answer inside \textbackslash boxed\{\} (e.g. \textbackslash boxed\{4\})
\end{promptbox}

\subsection{Real-World Transfer Evaluation Prompts}
\label{app_sub:realworld_prompts}

To evaluate model transfer performance on real-world physical activity videos (e.g., the RepCount / TransRAC benchmark), we test zero-shot, chain-of-thought (\texttt{cot}), and direct prompt conditions using the following exact templates:

\begin{promptbox}{Real-World Zero-Shot Prompt (\texttt{zero\_shot})}
How many times does the person perform the repetitive exercise in this video? Show your reasoning and put the final answer in \textbackslash boxed\{\}.
\end{promptbox}

\begin{promptbox}{Real-World Chain-of-Thought Prompt (\texttt{cot})}
How many times does the person perform the repetitive exercise in this video? First, describe what exercise is being performed. Then, enumerate each individual repetition you observe, noting approximately when each one occurs. Finally, provide your total count. Put the final answer in \textbackslash boxed\{\}.
\end{promptbox}

\begin{promptbox}{Real-World Direct Answer Prompt (\texttt{direct})}
Count the number of exercise repetitions in this video. Reply with only a number.
\end{promptbox}

\subsection{Output Parsing and Evaluation}
\label{app_sub:output_parsing}

We parse the final integer from each response, treating an unparseable answer as incorrect. Trace extraction is described in Appendix~\ref{app:trace_evaluation_and_validation}. Briefly, only explicit timestamps in the visible response are treated as reported events; ordinal descriptions and final counts alone do not yield timestamps. Consequently, the trace metrics quantify the evidence externalized by each prompting condition. Direct-answer outputs are scored for exact match, and any trace score is based solely on timestamps voluntarily included in the response.

\subsection{Cross-Model Prompting Results}
\label{app_sub:cross_model_prompting_results}

Table~\ref{tab:app_cross_model_prompting} places all available Qwen3-VL-8B and InternVL3.5-8B prompt conditions beside the corresponding Gemini reference conditions. The render sets differ slightly but share the Bounce Ball task and parameterization.

\begin{table*}[t]
\centering
\small
\begin{tabular}{llrrrrrrr}
\toprule
\textbf{Model} & \textbf{Prompt} & \textbf{EM (\%)} & \textbf{VOR} & \textbf{P (\%)} & \textbf{R (\%)} & \textbf{Trace \(F_1\) (\%)} & \textbf{ACR (\%)} & \textbf{RFR (\%)} \\
\midrule
Gemini 3.6 Flash & Direct & 20.4 & 1.16 & 45.6 & 40.2 & 38.5 & 7.5 & 3.2 \\
Gemini 3.6 Flash & Structured trace & 19.6 & 0.71 & 59.2 & 31.5 & 36.6 & 3.9 & 3.6 \\
Gemini 3.6 Flash & Multi-turn & 20.3 & 0.40 & 15.9 & 9.5 & 10.1 & 18.2 & 0.6 \\
Gemini 3.6 Flash & Thinking / CoT & 19.5 & 2.13 & 33.8 & 48.5 & 36.9 & 14.3 & 2.1 \\
Gemini 3.6 Flash & Role prompt & 19.3 & 1.52 & 37.2 & 44.9 & 37.6 & 10.8 & 2.1 \\
\midrule
Qwen3-VL-8B & Direct & 12.4 & 28.76 & 16.8 & 17.5 & 13.0 & 4.9 & 1.7 \\
Qwen3-VL-8B & CoT & 8.2 & 18.69 & 11.1 & 30.5 & 11.6 & 2.6 & 0.8 \\
Qwen3-VL-8B & Multi-turn & 13.5 & 0.26 & 23.8 & 11.5 & 13.4 & 4.7 & 1.7 \\
Qwen3-VL-8B-Thinking & CoT & 10.2 & 0.17 & 10.9 & 4.3 & 5.4 & 5.6 & 0.4 \\
\midrule
InternVL3.5-8B & Direct & 9.0 & 0.00 & 0.0 & 0.0 & 0.0 & 9.9 & 0.0 \\
InternVL3.5-8B & CoT & 9.8 & 0.00 & 0.0 & 0.0 & 0.0 & 9.0 & 0.0 \\
InternVL3.5-8B & Multi-turn & 9.8 & 0.00 & 0.0 & 0.0 & 0.0 & 10.7 & 0.0 \\
\bottomrule
\end{tabular}
\caption{\textbf{Task-matched cross-model prompting and trace summary on Bounce Ball.} All model profiles evaluate 730 trials per domain. All trace metrics use frequency-relative timestamp matching and score only explicit, parser-recognized timestamps.}
\label{tab:app_cross_model_prompting}
\end{table*}

Gemini's prompt variants have similar final-answer accuracy but substantially different reported-evidence profiles. The available Qwen conditions exhibit a related separation. Multi-turn produces the highest final-answer exact match (\(13.5\%\)) and trace \(F_1\) (\(13.4\%\)) among the listed Qwen variants, whereas Direct and CoT exhibit substantial over-reporting and low precision. InternVL produces nonzero final answers in every condition but no extractable timestamps. These results distinguish the evidence expressed under each prompt from any unobserved internal reasoning process. In particular, the table shows that changing the requested reasoning format can change the visible trace substantially without establishing a corresponding change in temporal capability.

\subsection{Frame-Sampling Density Sweeps}
\label{app_sub:sampling_density}
\label{app:visual_access_interventions}

All visual-access interventions evaluate Bounce Ball, Blinking, and State Machine visual tasks using the same rendered videos and seeds as the native-video baseline. The sampling sweep changes only the density of supplied frames. The main paper reports the corresponding Gemini results. This appendix provides the complete cross-domain Gemini and open-model visual-access evidence below.

\subsection{Event-Centered Keyframe Extraction}
\label{app_sub:keyframe_evidence}

For each true event at time \(t_i\), the keyframe condition provides the frame triplet
\[
\mathcal{K}_i=\{I(t_i-0.1\,\mathrm{s}), I(t_i), I(t_i+0.1\,\mathrm{s})\}.
\]
This is an upper-bound visual-access probe because one frame group is provided for every true event. It removes the need to search the full video for event locations while preserving the requirement to retain and aggregate event evidence.

\subsection{Complete Gemini Intervention Analyses}
\label{app_sub:complete_gemini_intervention_surfaces}

\begin{figure*}[htbp]
    \centering
    \includegraphics[width=0.98\textwidth]{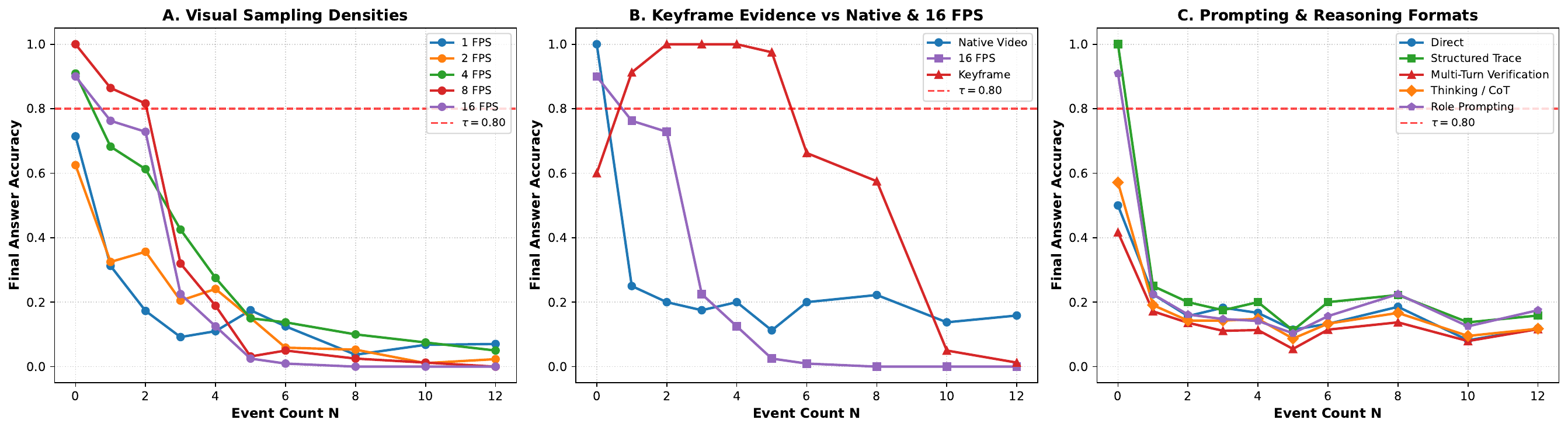}
    \caption{\textbf{Bounce Ball interventions.} Final-answer exact match averaged over frequencies across visual sampling densities, keyframe evidence, and prompting strategies.}
    \label{fig:intervention_analysis_bounce_ball}
\end{figure*}

\begin{figure*}[htbp]
    \centering
    \includegraphics[width=0.98\textwidth]{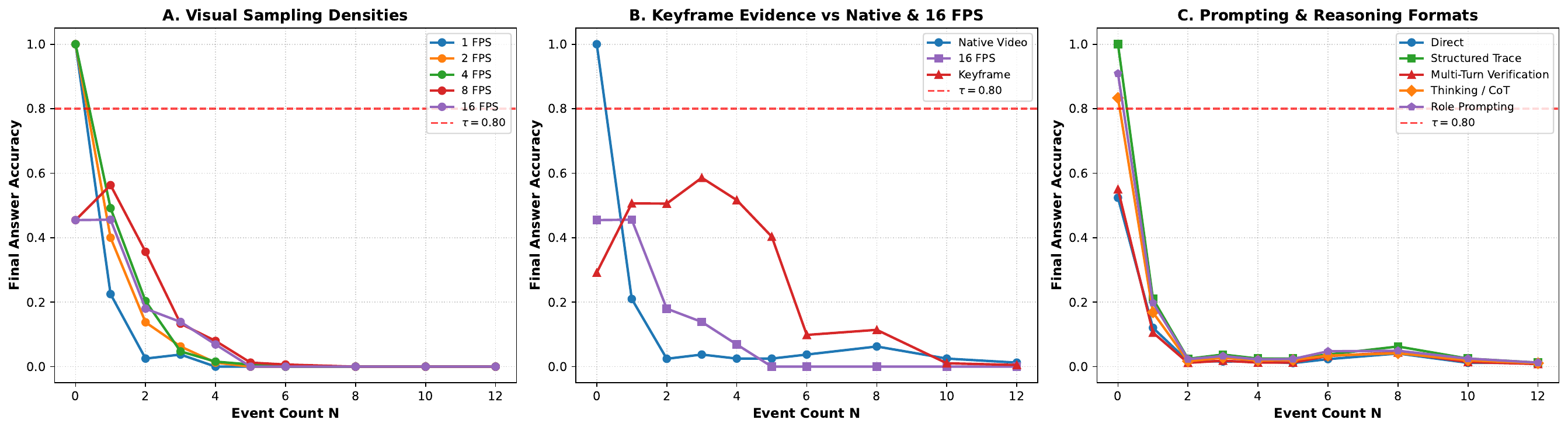}
    \caption{\textbf{Blinking interventions.} Final-answer exact match averaged over frequencies across visual sampling densities, keyframe evidence, and prompting strategies.}
    \label{fig:intervention_analysis_blinking}
\end{figure*}

\begin{figure*}[htbp]
    \centering
    \includegraphics[width=0.98\textwidth]{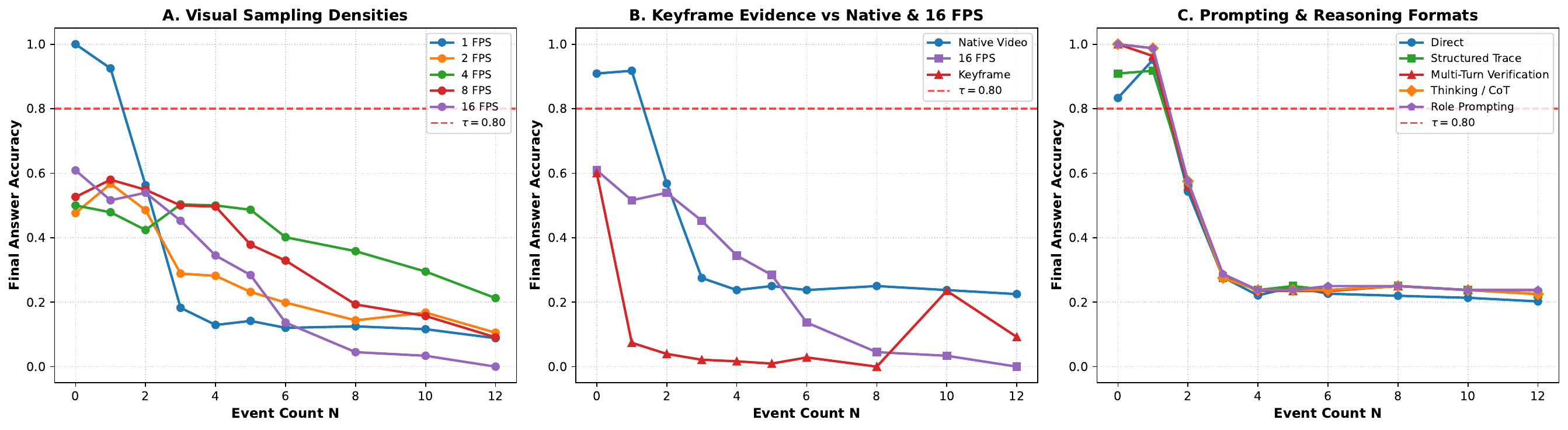}
    \caption{\textbf{State Machine interventions.} Final-answer exact match averaged over frequencies across visual sampling densities, keyframe evidence, and prompting strategies.}
    \label{fig:intervention_analysis_state_machine}
\end{figure*}

Figures~\ref{fig:intervention_analysis_bounce_ball}, \ref{fig:intervention_analysis_blinking}, and \ref{fig:intervention_analysis_state_machine} present the 3-panel intervention analysis across visual sampling densities, keyframe evidence, and prompting formats for all three benchmark domains.

\begin{figure*}[p]
    \centering
    \includegraphics[width=0.98\textwidth]{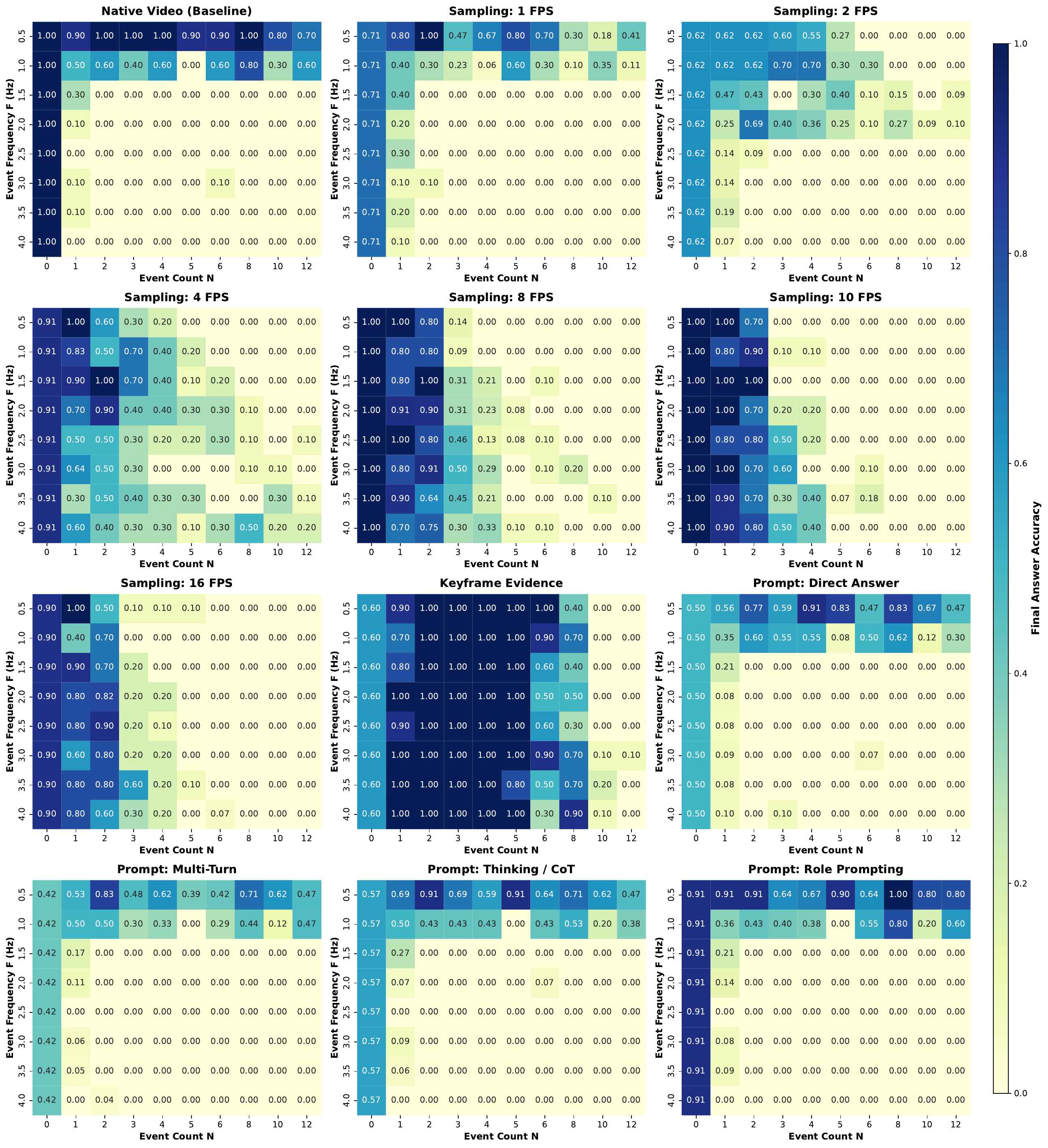}
    \caption{\textbf{Complete Gemini visual-access intervention surfaces on Bounce Ball.} Final-answer exact-match accuracy for native video, supplied frame-sampling densities, and event-centered keyframes.}
    \label{fig:full_intervention_heatmaps_bounce_ball}
\end{figure*}

\begin{figure*}[p]
    \centering
    \includegraphics[width=0.98\textwidth]{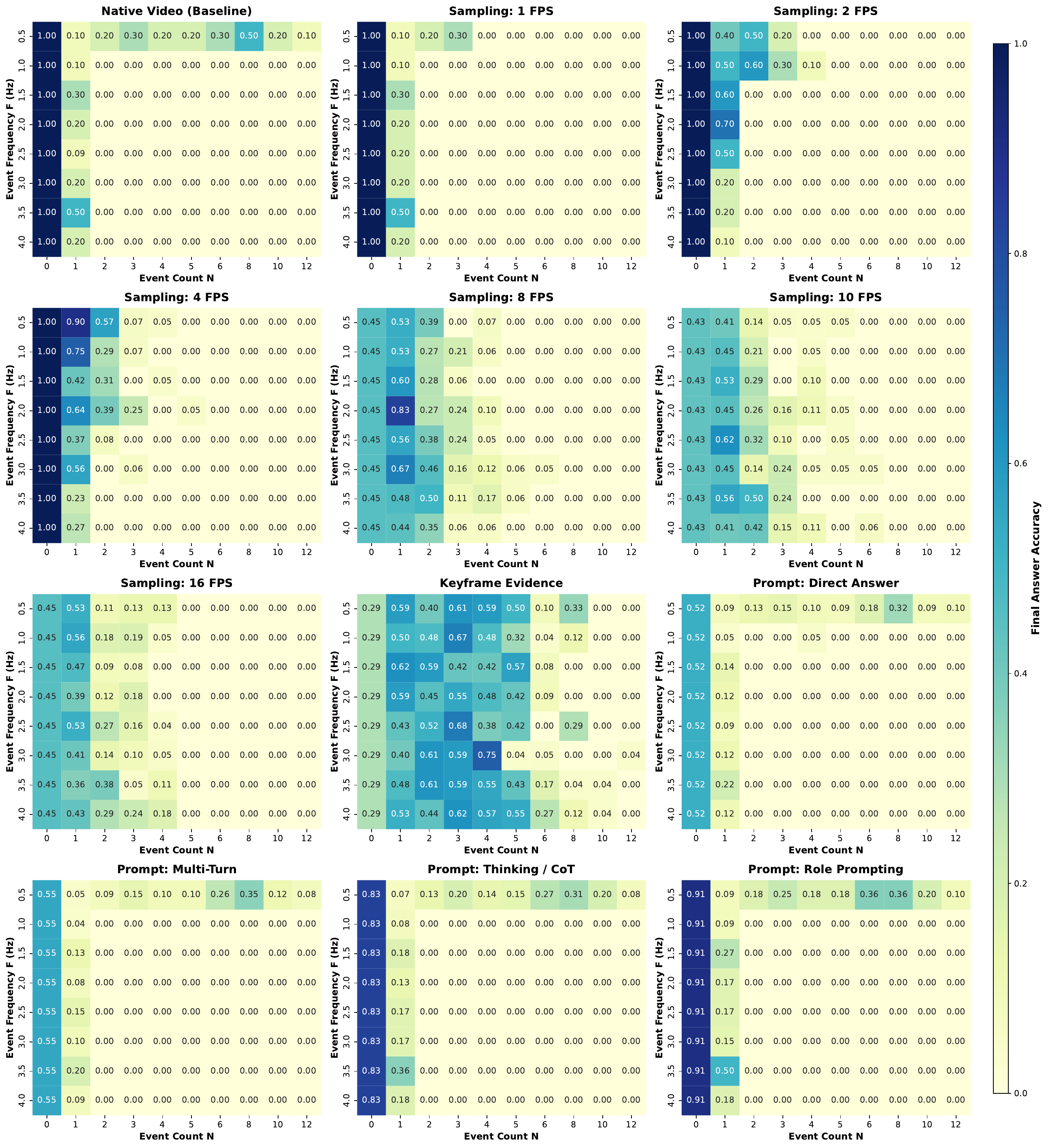}
    \caption{\textbf{Complete Gemini visual-access intervention surfaces on Blinking.} Final-answer exact-match accuracy for native video, supplied frame-sampling densities, and event-centered keyframes.}
    \label{fig:full_intervention_heatmaps_blinking}
\end{figure*}

\begin{figure*}[p]
    \centering
    \includegraphics[width=0.98\textwidth]{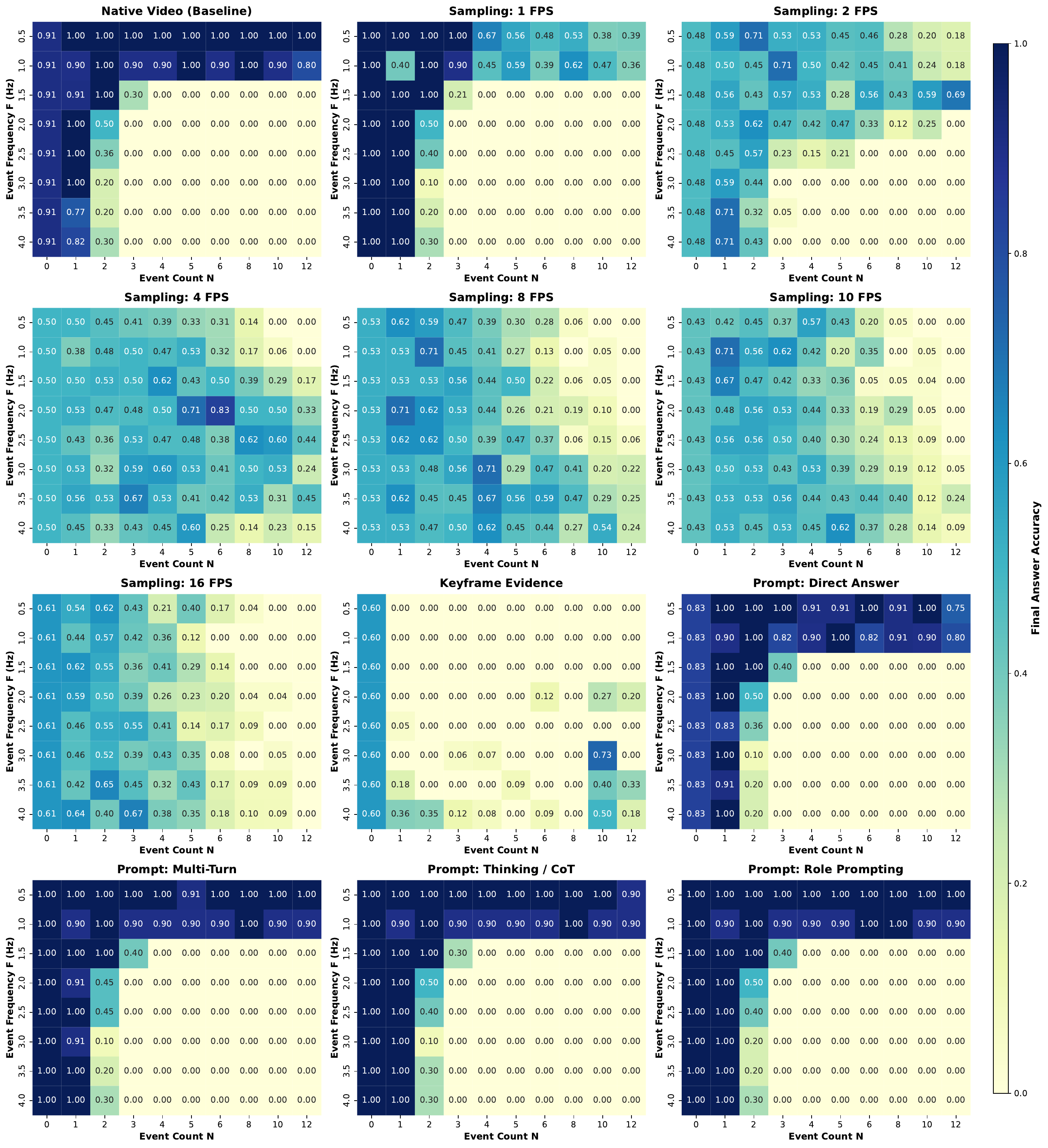}
    \caption{\textbf{Complete Gemini visual-access intervention surfaces on State Machine.} Final-answer exact-match accuracy for native video, supplied frame-sampling densities, and event-centered keyframes.}
    \label{fig:full_intervention_heatmaps_state_machine}
\end{figure*}

Full cell-by-cell 12-panel capability surface heatmaps for Bounce Ball (Figure~\ref{fig:full_intervention_heatmaps_bounce_ball}), Blinking (Figure~\ref{fig:full_intervention_heatmaps_blinking}), and State Machine (Figure~\ref{fig:full_intervention_heatmaps_state_machine}) are detailed in the heatmap section preceding the qualitative examples. Denser regular sampling improves a limited low-count portion of the surface, whereas event-centered keyframes extend accuracy on discrete physical contact domains (Bounce Ball exact match increases from $19.5\%$ baseline up to $68.6\%$). However, on the State Machine domain, oracle keyframe evidence causes a severe performance collapse for Gemini 3.6 Flash (exact match drops from $36.8\%$ baseline and $40.7\%$ under 4 FPS down to $6.3\%$). Because state transitions represent categorical mode shifts requiring continuous temporal context to verify state persistence, stripping non-event frames removes the background state history necessary for sequential state tracking.

\subsection{Open-Model Keyframe and Sampling Interventions}
\label{app_sub:qwen_visual_access_interventions}

On the State Machine domain, supplying event-centered keyframe evidence substantially improves open-model exact match. For Qwen3-VL-8B, exact match increases from $26.6\%$ under native video (800 trials) to $69.4\%$ under event-centered keyframes (720 trials), a gain of $+42.8$ percentage points. For InternVL3.5-8B, keyframe triplets improve exact match from $26.2\%$ to $32.1\%$ ($+5.9$ points).

\begin{figure}[h]
    \centering
    \includegraphics[width=\linewidth]{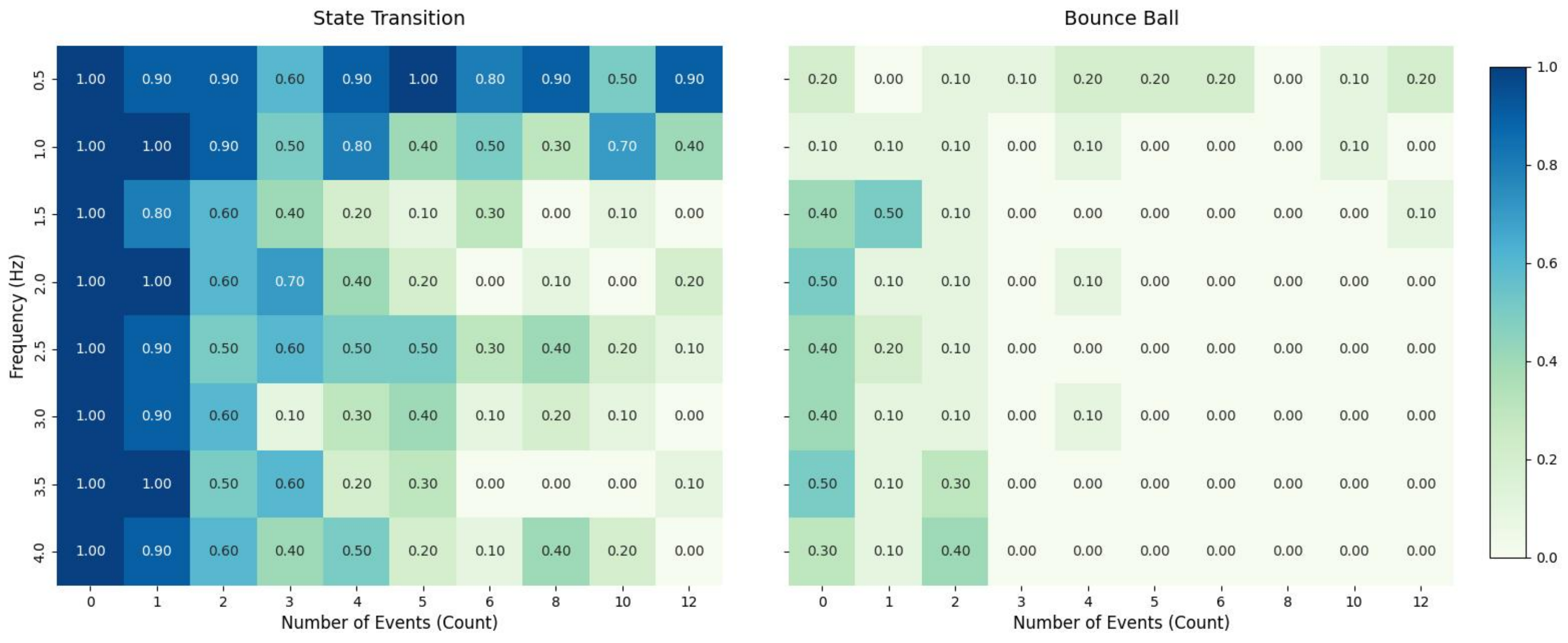}
    \caption{\textbf{Qwen3-VL-8B-Instruct under denser frame sampling (10 FPS).} Exact-match capability surfaces across temporal counting tasks under 10-FPS supplied-frame preprocessing. The figure provides an open-model counterpart to the visual-access interventions.}
    \label{fig:app_qwen_fps_sampling}
\end{figure}

\begin{figure}[h]
    \centering
    \includegraphics[width=\linewidth]{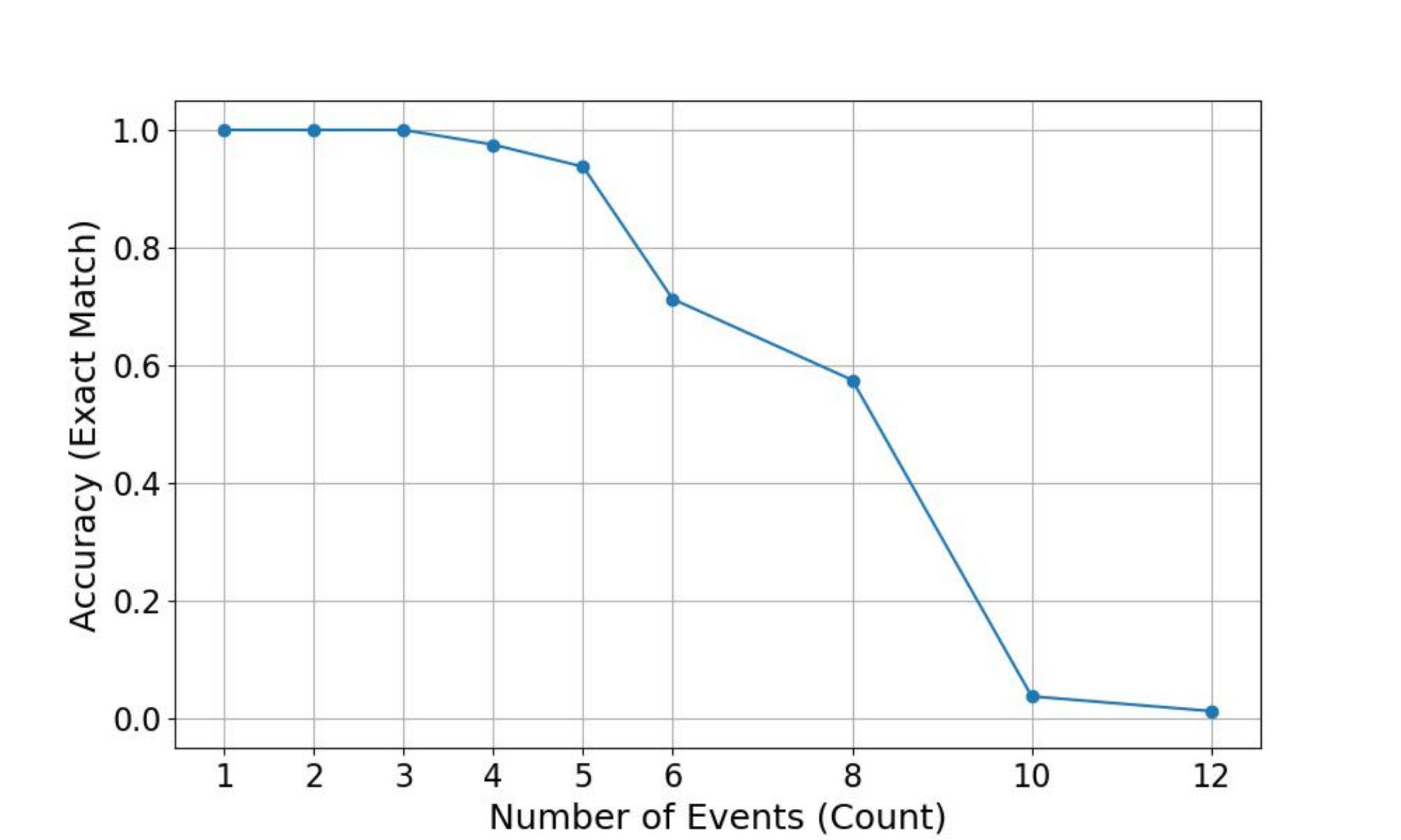}
    \caption{\textbf{Oracle keyframe evidence intervention analysis for Qwen3-VL-8B-Instruct.} Exact-match accuracy across event count and frequency when models are supplied with oracle event-centered keyframe triplets versus baseline video sampling.}
    \label{fig:app_keyframe_analysis}
\end{figure}

\subsection{GPT-5.6 Sol Frame-Sequence Interventions \& API Limitations}
\label{app_sub:gpt_sol_interventions}

Unlike native video models, GPT-5.6 Sol does not accept raw container video files directly. Instead, videos are manually sampled into ordered sequences of image frames before being passed to the API.

\begin{figure*}[p]
    \centering
    \includegraphics[width=0.98\textwidth]{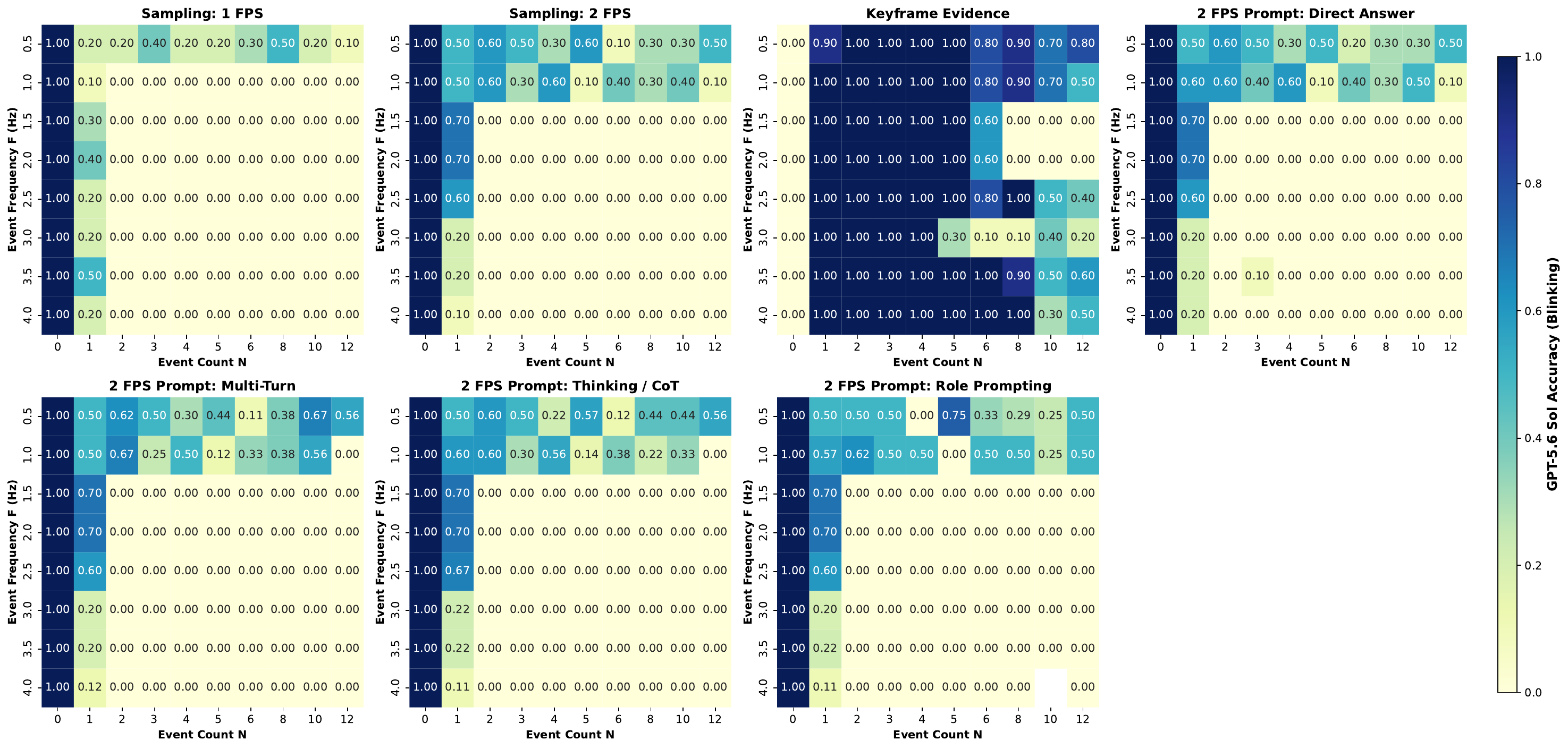}
    \caption{\textbf{GPT-5.6 Sol capability surfaces on Blinking.} Final-answer exact-match accuracy under 1--2 FPS sampling densities, oracle keyframes, and 2-FPS prompting interventions.}
    \label{fig:app_gpt_sol_heatmaps_blinking}
\end{figure*}

\begin{figure*}[p]
    \centering
    \includegraphics[width=0.98\textwidth]{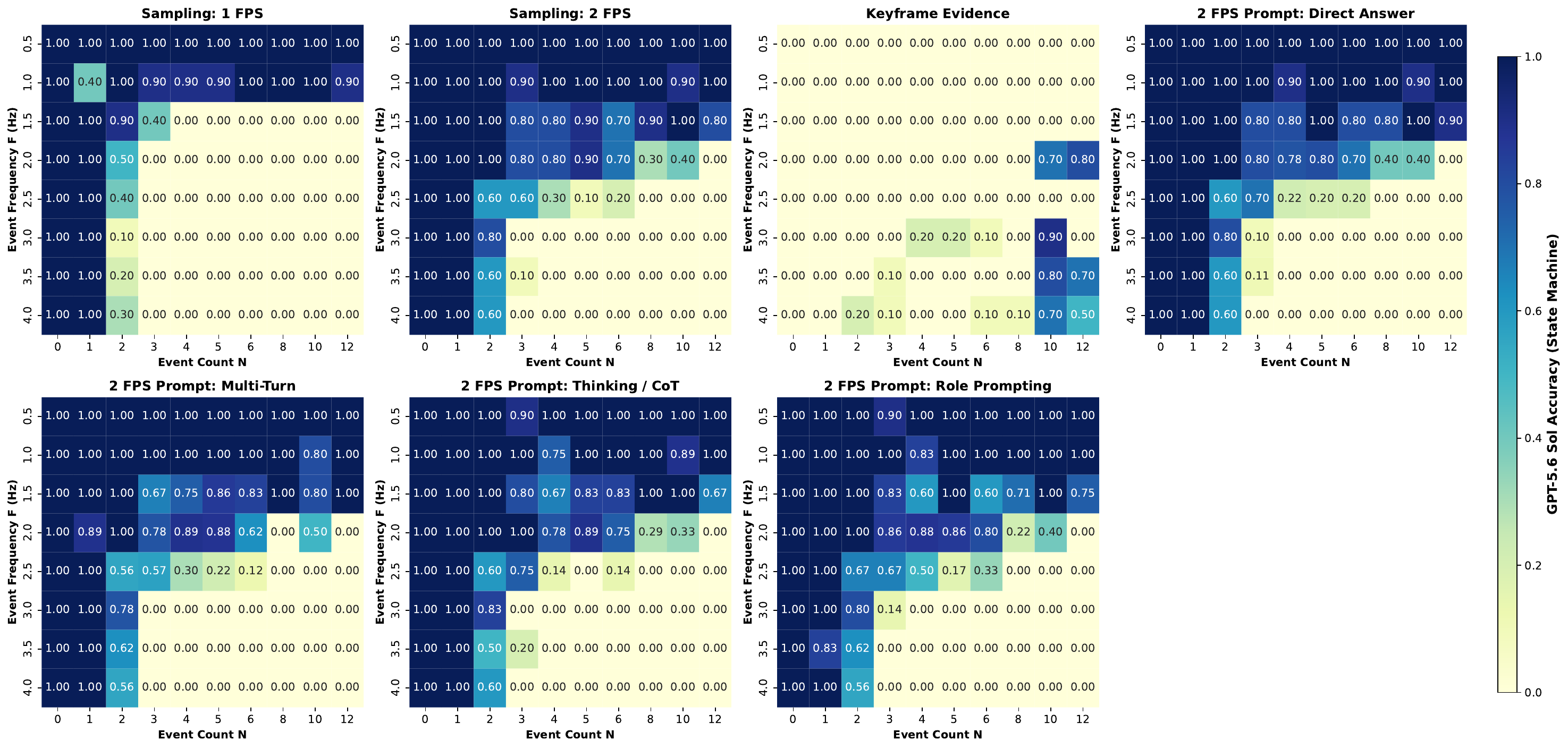}
    \caption{\textbf{GPT-5.6 Sol capability surfaces on State Machine.} Final-answer exact-match accuracy under 1--2 FPS sampling densities, oracle keyframes, and 2-FPS prompting interventions.}
    \label{fig:app_gpt_sol_heatmaps_state_machine}
\end{figure*}

\paragraph{API Frame Ceiling \& Token Context Limitations.} During multi-density sampling sweeps, GPT-5.6 Sol encountered strict API payload constraints. Specifically, the deployment environment enforces a maximum limit of 50 images per request (\texttt{BadRequestError: Too many images in request: 51, maximum allowed: 50}). Consequently, 24-second videos sampled at higher densities (4 FPS = 96 frames, 8 FPS = 192 frames, 10 FPS = 240 frames, 16 FPS = 384 frames) failed due to payload size and token context window limits. Evaluation was successfully completed for 1 FPS (24 frames $\le 50$), 2 FPS (48 frames $\le 50$), Oracle Keyframe Evidence ($\approx 3 N$ frames), and 2-FPS prompting interventions (\texttt{direct}, \texttt{structured\_trace}, \texttt{multi\_turn\_verification}, \texttt{thinking}, \texttt{role\_prompting}). Table~\ref{tab:app_gpt_sol_trace} provides the quantitative trace summary across protocols, and the corresponding capability surfaces are reported in Figure~\ref{fig:app_gpt_sol_heatmaps_blinking} (Blinking) and Figure~\ref{fig:app_gpt_sol_heatmaps_state_machine} (State Machine).

\begin{table*}[t]
\centering
\small
\begin{tabularx}{\textwidth}{Xrrrrrrr}
\toprule
\textbf{Model / Intervention Protocol} & \textbf{EM (\%)} & \textbf{VOR} & \textbf{P (\%)} & \textbf{R (\%)} & \textbf{Trace \(F_1\) (\%)} & \textbf{ACR (\%)} & \textbf{RFR (\%)} \\
\midrule
GPT-5.6 Sol (1 FPS Structured Trace) & 21.9 & 0.88 & 87.5 & 62.2 & 63.9 & 15.8 & 29.7 \\
GPT-5.6 Sol (2 FPS Structured Trace) & 34.9 & 0.86 & 86.4 & 60.4 & 60.7 & 21.9 & 21.4 \\
GPT-5.6 Sol (2 FPS Direct Answer) & 35.5 & 0.01 & 1.4 & 1.4 & 1.4 & 34.1 & 0.0 \\
GPT-5.6 Sol (2 FPS Multi-Turn) & 35.1 & 0.39 & 66.1 & 38.6 & 44.6 & 12.6 & 5.7 \\
GPT-5.6 Sol (2 FPS Thinking / CoT) & 29.9 & 0.05 & 12.8 & 4.5 & 5.6 & 27.3 & 0.4 \\
GPT-5.6 Sol (2 FPS Role Prompt) & 39.5 & 0.48 & 58.6 & 39.9 & 42.4 & 25.8 & 13.5 \\
GPT-5.6 Sol (Oracle Keyframes) & 42.5 & 1.18 & 81.2 & 79.6 & 71.1 & 20.4 & 34.9 \\
\bottomrule
\end{tabularx}
\caption{\textbf{GPT-5.6 Sol intervention trace evaluation summary.} Macro averages reported on non-errored task-matched evaluation records across sampling densities, oracle keyframes, and prompting formats.}
\label{tab:app_gpt_sol_trace}
\end{table*}

\paragraph{Keyframe Breakdown on State Transitions vs Flash Events.} On the Blinking task, supplying oracle keyframe triplets ($\pm 0.1\text{s}$ around event timestamps) elevates GPT-5.6 Sol exact match from $7.1\%$ (1 FPS) and $14.4\%$ (2 FPS) up to $76.4\%$, as isolated keyframe thumbnails directly capture the visual flash on-pulse. Conversely, on the State Machine task, oracle keyframes yield low exact match ($8.5\%$), whereas continuous 2-FPS frame sampling achieves GPT-5.6 Sol's highest performance ($55.5\%$--$58.8\%$ exact match). This discrepancy highlights a fundamental domain difference: while discrete flash events can be identified from isolated keyframes, tracking state machine transitions requires continuous intermediate visual context to verify color/state persistence and track cumulative state history. Stripping non-event frames removes the intermediate state evidence necessary to track sequential transitions across the video.

\subsection{Trace Extraction and Matching Algorithm}
\label{app_sub:trace_matching_algorithm}
\label{app:trace_evaluation_and_validation}

For every evaluation record, we retrieve the executable trace emitted
by the renderer for the corresponding video and deduplicate repeated
event entries in the log. Gemini is prompted to produce a structured
event ledger, from which we extract explicit minute--second entries.
For the open models, whose responses are free-form, we extract
minute--second and decimal-second expressions preceded by temporal
cues such as ``at'', ``from'', or ``frame''. Mentions within 0.1
seconds are deduplicated. Ordinal descriptions and final counts alone
do not create reported events. This conservative procedure avoids
assigning timestamps that the model did not provide, while making trace
metrics conditional on the temporal evidence externalized in the
response.

For a ground-truth trace with \(N\) events and a model report with
\(M\) events, we form candidate pairs whose timestamps differ by at
most \(\delta(F)=1/(2F)\). This rate-relative tolerance ranges from
\(1.0\) second at \(0.5\) Hz to \(0.125\) second at \(4.0\) Hz and
keeps adjacent true-event windows disjoint. We choose a one-to-one
matching that maximizes the number of aligned pairs \(K\), breaking
ties by the smallest total timestamp deviation.

An unmatched true event contributes to a miss and an unmatched
reported event contributes to an unsupported report. The reported
trace then yields precision \(K/M\), recall \(K/N\), trace \(F_1\),
and the Visual Observation Ratio \(M/N\). When \(M=0\), precision and
trace \(F_1\) are defined as zero. A correct final count with
trace \(F_1<0.80\) contributes to ACR, while an incorrect final count
with trace \(F_1\geq0.80\) contributes to RFR.

\subsection{Aggregate Trace Error Summary}
\label{app_sub:trace_error_cases}

Table~\ref{tab:app_error_taxonomy} defines the task-general taxonomy
used for the case inventory. State-specific discrepancies are treated
as instances of misidentified events rather than as a separate type.

\begin{table*}[t]
\centering
\small
\begin{tabular}{p{0.26\textwidth}p{0.68\textwidth}}
\toprule
\textbf{Type} & \textbf{Definition} \\
\midrule
Correct match & A reported event aligns one-to-one with a true event within the timestamp tolerance and has the correct available event attribute. \\
Missed event & A true event has no aligned reported event. \\
Merged or collapsed events & The visible response explicitly combines two or more true events into one report. This type is assigned only when the response supports that interpretation. \\
Hallucinated event & A reported event has no aligned true event. \\
Temporally displaced event & A report corresponds to the appropriate event in the sequence but falls outside the timestamp tolerance. \\
Misidentified event & A timestamp aligns, but the reported event attribute is incorrect, such as the contacted wall, blink phase, or resulting state. \\
Wrong accumulation & The reported event sequence is sufficiently faithful with trace \(F_1\geq0.80\), but the final count is incorrect. \\
\bottomrule
\end{tabular}
\caption{\textbf{Task-general trace-error taxonomy.} The first six types concern individual reported or true events. Wrong accumulation concerns the relationship between an otherwise faithful report and the final answer.}
\label{tab:app_error_taxonomy}
\end{table*}

Table~\ref{tab:app_trace_cases} aggregates these categories over all
positive-count videos for each model. Matched, missed, hallucinated,
displaced, and misidentified entries are event-level counts. Wrong
accumulation is a video-level count. We report zero merged or
collapsed events because no visible response was manually verified to
combine multiple true events into one report. Gemini has the largest
number of matched events, while the open-model rows show either many
unmatched timestamp entries or no parser-recognized temporal report.
The latter should not be read as absence of visual processing. It
means that the visible answer did not provide timestamped evidence
under the extraction rule.

\begin{table*}[t]
\centering
\small
\begin{tabular}{lrrrrrrr}
\toprule
\textbf{Error type} & \textbf{Gemini} & \textbf{Qwen} & \textbf{Qwen} & \textbf{Qwen} & \textbf{InternVL} & \textbf{InternVL} & \textbf{InternVL} \\
& \textbf{3.6 Flash} & \textbf{3-VL-8B} & \textbf{3-VL-32B} & \textbf{3-VL-235B} & \textbf{3.5-8B} & \textbf{3.5-30B-A3B} & \textbf{3.5-38B} \\
\midrule
True events & 12,240 & 12,223 & 12,223 & 12,223 & 12,223 & 12,223 & 12,223 \\
Matched events & 3,907 & 2,458 & 1,678 & 1,130 & 0 & 0 & 0 \\
Missed events & 8,333 & 9,765 & 10,545 & 11,093 & 12,223 & 12,223 & 12,223 \\
Merged events & 0 & 0 & 0 & 0 & 0 & 0 & 0 \\
Hallucinated events & 2,096 & 88,506 & 2,348 & 931 & 0 & 0 & 0 \\
Displaced events & 448 & 245 & 402 & 142 & 0 & 0 & 0 \\
Misidentified events & 1,556 & 737 & 518 & 200 & 0 & 0 & 0 \\
Wrong accumulation & 50 & 34 & 34 & 23 & 0 & 0 & 0 \\
Videos & 2,160 & 2,160 & 2,160 & 2,160 & 2,160 & 2,160 & 2,160 \\
\bottomrule
\end{tabular}
\caption{\textbf{Aggregate trace-error counts by model.} Counts aggregate all positive-count videos across Bounce Ball, Blinking, and State Machine. Matched, missed, hallucinated, displaced, and misidentified rows count events. Wrong accumulation counts videos with an incorrect final answer and trace \(F_1\geq0.80\). A displaced event is an order-aligned timestamp outside the tolerance. A misidentified event has an aligned timestamp but an incorrect available event attribute. Merged events require explicit visible evidence and are not inferred from missing timestamps. Gemini and open-model sets contain 12,240 and 12,223 true events, respectively.}
\label{tab:app_trace_cases}
\end{table*}

Table~\ref{tab:tolerance_sensitivity} reports the sensitivity of the
trace metrics to the timestamp-matching tolerance. Final-answer exact
match and VOR are unchanged because they do not depend on pairing.
Precision, recall, and trace \(F_1\) increase as the tolerance widens,
as expected. The selected rate-relative tolerance lies between the
fixed settings and preserves non-overlapping matching windows for
adjacent events at every tested frequency.

\begin{table*}[t]
\centering
\small
\begin{tabular}{lrrrrrr}
\toprule
\textbf{Matching Tolerance Strategy} & \textbf{Acc (\%)} & \textbf{VOR} & \textbf{Precision (\%)} & \textbf{Recall (\%)} & \textbf{Trace $F_1$ (\%)} & \textbf{ACR (\%)} \\
\midrule
Fixed \(\delta = 1.0\,\text{s}\) & 21.1 & 0.68 & 75.6 & 41.2 & 47.6 & 1.9 \\
Fixed \(\delta = 0.5\,\text{s}\) & 21.1 & 0.68 & 67.8 & 35.2 & 41.3 & 6.4 \\
Fixed \(\delta = 0.25\,\text{s}\) & 21.1 & 0.68 & 49.7 & 21.7 & 27.2 & 14.2 \\
Rate-Relative \(\delta = \frac{1}{2F}\) & 21.1 & 0.68 & 58.0 & 31.2 & 36.4 & 5.7 \\
\bottomrule
\end{tabular}
\caption{\textbf{Trace-metric sensitivity to timestamp tolerance for the Gemini 3.6 Flash baseline.} Final EM averages the full grid; VOR, precision, recall, trace $F_1$, and ACR average positive-event trials only. The rate-relative tolerance $\delta(F)=1/(2F)$ bounds matching within each event's unique inter-event half-period.}
\label{tab:tolerance_sensitivity}
\end{table*}

\section{Computational Cost Analysis}
\label{app_sub:cost_analysis}

To ensure full transparency and reproducible benchmarking standards, we record financial API costs incurred across all evaluation protocols. In total, across all 5 core experimental sweeps and real-world transfer evaluations across all three visual domains, the benchmark evaluation processed 37,924 individual evaluation trials, incurring a total financial API evaluation cost of \$3,661.78 USD.

\begin{table*}[htbp]
\centering
\small
\begin{tabularx}{\textwidth}{Xrrrrr}
\toprule
\textbf{Experiment / Protocol} & \textbf{Trials} & \textbf{Input Tokens} & \textbf{Output Tokens} & \textbf{Total Tokens} & \textbf{Cost (\$USD)} \\
\midrule
Exp 1: Baseline $N \times F$ Matrix Sweep & 2,203 & 3,659,556 & 1,803,571 & 5,463,127 & \$19.02 \\
Exp 2: Frame Density Interventions (1--16 FPS) & 20,102 & 2,438,029,486 & 15,172,075 & 2,453,201,561 & \$3,426.86 \\
Exp 3: Oracle Keyframe Evidence Interventions & 3,449 & 45,757,799 & 3,050,018 & 48,807,817 & \$89.47 \\
Exp 4: Prompting Strategies \& Thinking Modes & 11,714 & 18,905,824 & 10,567,098 & 29,472,922 & \$107.61 \\
Real-World Transfer (RepCount / TransRAC) & 456 & 8,014,360 & 906,931 & 8,921,291 & \$18.82 \\
\midrule
\textbf{Total Benchmark Evaluation} & \textbf{37,924} & \textbf{2,514,367,025} & \textbf{31,499,693} & \textbf{2,545,866,718} & \textbf{\$3,661.78} \\
\bottomrule
\end{tabularx}
\caption{\textbf{Comprehensive benchmark evaluation cost and token usage breakdown across all three visual domains.} Summarizes total evaluation trials, input/output token counts, and financial API costs across all 5 core experiments and real-world transfer evaluation.}
\label{tab:app_cost_summary}
\end{table*}

Table~\ref{tab:app_cost_summary} presents the high-level cost breakdown across experiments. The vast majority of financial expenditure (\$3,426.86 USD, or 93.6\% of total cost) was consumed by Experiment~2 (Frame Sampling Density Interventions), as high sampling densities (8--16 FPS) convert 24-second videos into long sequences of image patches, creating massive context window loads across all three visual domains.

\begin{table*}[htbp]
\centering
\small
\begin{tabularx}{0.75\textwidth}{Xrrrr}
\toprule
\textbf{Frame Density} & \textbf{Trials} & \textbf{Input Tokens} & \textbf{Output Tokens} & \textbf{Cost (\$USD)} \\
\midrule
Native Video & 2,203 & 3,659,556 & 1,803,571 & \$19.02 \\
1 FPS & 2,862 & 58,130,830 & 2,004,410 & \$95.80 \\
2 FPS & 2,933 & 115,870,930 & 2,206,109 & \$167.95 \\
4 FPS & 3,428 & 225,535,430 & 2,479,566 & \$313.27 \\
8 FPS & 3,453 & 452,188,230 & 2,633,080 & \$662.41 \\
10 FPS & 3,554 & 589,062,945 & 2,692,969 & \$839.01 \\
16 FPS & 3,872 & 997,241,121 & 3,155,941 & \$1,348.43 \\
\bottomrule
\end{tabularx}
\caption{\textbf{Cost breakdown by frame sampling density across all tasks (Experiment 2).} Demonstrates how frame sampling rate converts video inputs into large image-patch token sequences, scaling input processing costs.}
\label{tab:app_cost_density}
\end{table*}

\begin{table*}[htbp]
\centering
\small
\begin{tabularx}{0.75\textwidth}{Xrrrr}
\toprule
\textbf{Visual Domain / Task} & \textbf{Trials} & \textbf{Input Tokens} & \textbf{Output Tokens} & \textbf{Cost (\$USD)} \\
\midrule
\texttt{state\_machine} & 13,141 & 910,900,516 & 9,798,474 & \$1,249.46 \\
\texttt{bounce\_ball} & 10,363 & 778,500,899 & 10,729,970 & \$1,247.40 \\
\texttt{blinking} & 13,964 & 816,951,250 & 10,064,318 & \$1,146.10 \\
\texttt{repcount} (Real-World) & 456 & 8,014,360 & 906,931 & \$18.82 \\
\midrule
\textbf{Total} & \textbf{37,924} & \textbf{2,514,367,025} & \textbf{31,499,693} & \textbf{\$3,661.78} \\
\bottomrule
\end{tabularx}
\caption{\textbf{Evaluation cost and token usage breakdown by visual domain.} Summarizes computational allocation across synthetic video domains and real-world transfer datasets.}
\label{tab:app_cost_domain}
\end{table*}

As detailed in Table~\ref{tab:app_cost_density}, increasing frame sampling rates from native video (1 FPS equivalent, \$19.02) up to 16 FPS (\$1,348.43) scales input processing costs exponentially. Importantly, despite supplying $16\times$ more visual frames and spending over \$3,420 USD on multi-frame density interventions across domains, the fundamental Low-Frequency Trap failure boundary remains completely intact. While extra visual frames provide minor bumps in final-integer count accuracy on \texttt{bounce\_ball} (from 19.6\% to 29.3\%), the faithful step-by-step trace agreement remains near zero (3.7\%), and models fail completely at high frequencies ($F \ge 2.0\text{ Hz}$). Thus, supplying higher-density visual inputs merely inflates evaluation costs exponentially without resolving the underlying architectural VLM bottleneck in temporal tracking and event bookkeeping. Table~\ref{tab:app_cost_domain} reports the domain-wise resource allocation across \texttt{state\_machine} (\$1,249.46 USD), \texttt{bounce\_ball} (\$1,247.40 USD), and \texttt{blinking} (\$1,146.10 USD).

\section{Extended Related Work}
\label{app:extended_related_work}

This section situates the paper against the work most directly relevant to controlled video evaluation and trace-grounded diagnosis.

\subsection{Video-Language Benchmarks and Controlled Evaluation}
Video-MME, TempCompass, HourVideo, EgoSchema, MVBench, LongVideoBench, Mementos, VideoNIAH, VideoCogQA, and Video-MMLU evaluate complementary aspects of video understanding \cite{fu2025videomme,liu2024tempcompass,chandrasegaran2024hourvideo,mangalam2023egoschemadiagnosticbenchmarklongform,li2024mvbench,wu2024longvideobenchbenchmarklongcontextinterleaved,wang2024mementos,videoniah,videocogqa,song2025videommlumassivemultidisciplinelecture}. VideoReasonBench focuses more specifically on multi-step reasoning over videos that contain partially observed state changes \cite{liu2025videoreasonbench}. These resources are complementary to our study, but their examples are fixed. Event count, event rate, duration, and visual complexity are not independently swept. In contrast, our generator changes count and frequency while holding the event semantics and rendering family fixed, which makes the resulting capability boundary directly interpretable.

Controlled visual benchmarks provide a useful precedent for this design. CLEVR and recent physics-oriented probes use controlled scenes to isolate compositional or physical-reasoning factors \cite{johnson2017clevr,chow2025physbench,xiang2025seephys,qiu2025phybench}. We extend that diagnostic philosophy to temporal event bookkeeping and then test the controlled finding on natural repeated-event videos.

\subsection{Programmatic Video and Intermediate Traces}
Recent controllable-video benchmarks use fully scripted clips and scalable task difficulty to stress-test multimodal reasoning. Our contribution is orthogonal to this controlled generation. Every video in our study is paired with a renderer-produced event schedule, which lets us score a model-reported event sequence rather than only its final answer.

Visual Reasoning Tracer is the closest precedent for evaluating intermediate visual reasoning traces. It assesses object-level traces in images against ground-truth visual annotations \cite{yuan2025visual}. Our setting differs in both modality and supervision. We evaluate temporal event sequences in video, and the reference trace is generated directly by the renderer. This permits timestamp-aware matching and separates missed or hallucinated events from errors in final count aggregation.

Broader work on reasoning evaluation motivates treating a final answer or a long rationale with care. Self-consistency and STaR improve or bootstrap language-model reasoning \cite{wang2022self,zelikman2022star}, while recent stress tests expose failures that emerge as reasoning tasks become more compositional or difficult \cite{shojaee2025illusionthinkingunderstandingstrengths,sun2025omegallmsreasonoutside}. In multimodal settings, EMMA, RCI, and Game-RL study reasoning capability, visual-information requirements, or verifiable task generation \cite{hao2025emma,agarwal2025rciscoreevaluatingglobal,tong2025gamerlsynthesizingmultimodalverifiable}. These are useful context, but none supplies a renderer-generated temporal event schedule against which a video model's reported event sequence can be aligned.

\section{Qualitative Trace Examples and Taxonomy Profiling}
\label{app:qualitative_examples}

This appendix presents representative model responses alongside the corresponding rendered key-event frame strips and executable ground-truth traces across all 6 diagnostic taxonomy categories. To ensure clear visual resolution while fitting comfortably within publication page height limits, each 5-sample category is presented across two page-optimized full-width figure panels: \textbf{Part 1} (Cases 1–3) and \textbf{Part 2} (Cases 4–5).

\subsection{Case-Selection Protocol}
\label{app_sub:qualitative_case_selection}

Qualitative cases were sampled from Gemini 3.6 Flash structured-trace predictions across all three synthetic benchmark domains (\texttt{bounce\_ball}, \texttt{blinking}, and \texttt{state\_machine}). For every sample card:
\begin{enumerate}
    \item \textbf{Event Frame Strips:} Every ground-truth event timestamp $t_e \in \text{GT\_events}$ is captured at its exact physical occurrence timestamp (highlighted with a green event boundary badge and timestamp label), complemented by context frames placed in start/end intervals.
    \item \textbf{Executable vs Reported Alignment:} Model-reported timestamps in the step-by-step event ledger are aligned against ground-truth event timestamps using a rate-relative matching tolerance ($\tau_{\text{match}} = 1.0\text{s}$).
    \item \textbf{Taxonomy Classification:} Samples are assigned to mutually exclusive failure categories based on final integer exact match and trace precision/recall/$F_1$.
\end{enumerate}

\subsection{Faithful Event Recovery (Correct Matches)}
\label{app_sub:qualitative_faithful_recovery}

\begin{figure*}[p]
    \centering
    \includegraphics[width=0.98\textwidth]{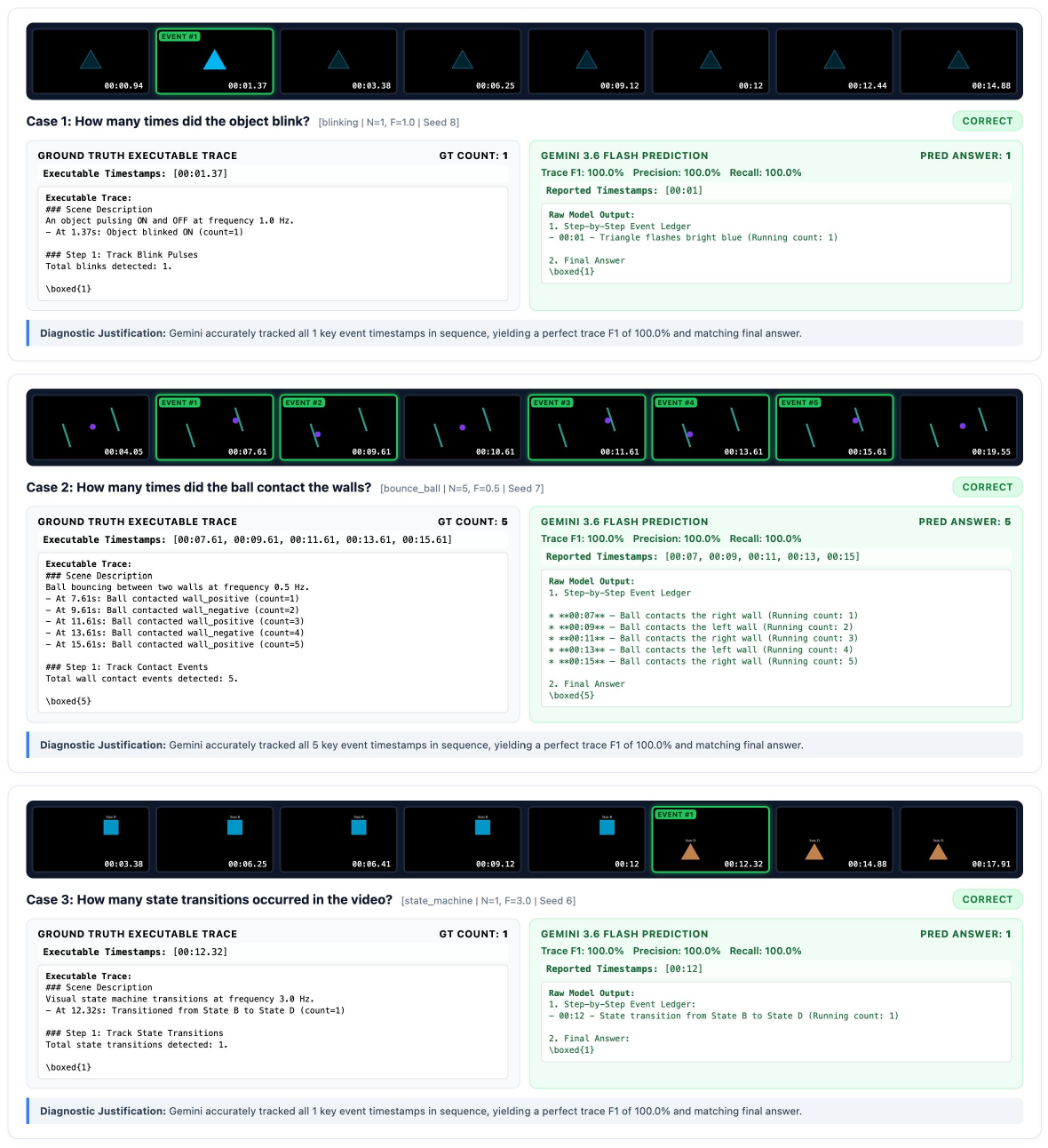}
    \caption{\textbf{Faithful Event Recovery (Part 1: Cases 1–3):} Gemini 3.6 Flash accurately tracks all key event timestamps and running counts on \texttt{blinking}, \texttt{bounce\_ball}, and \texttt{state\_machine}, achieving 100\% Trace $F_1$ and matching final integer counts.}
    \label{fig:app_qualitative_correct_part1}
\end{figure*}

\begin{figure*}[p]
    \centering
    \includegraphics[width=0.98\textwidth]{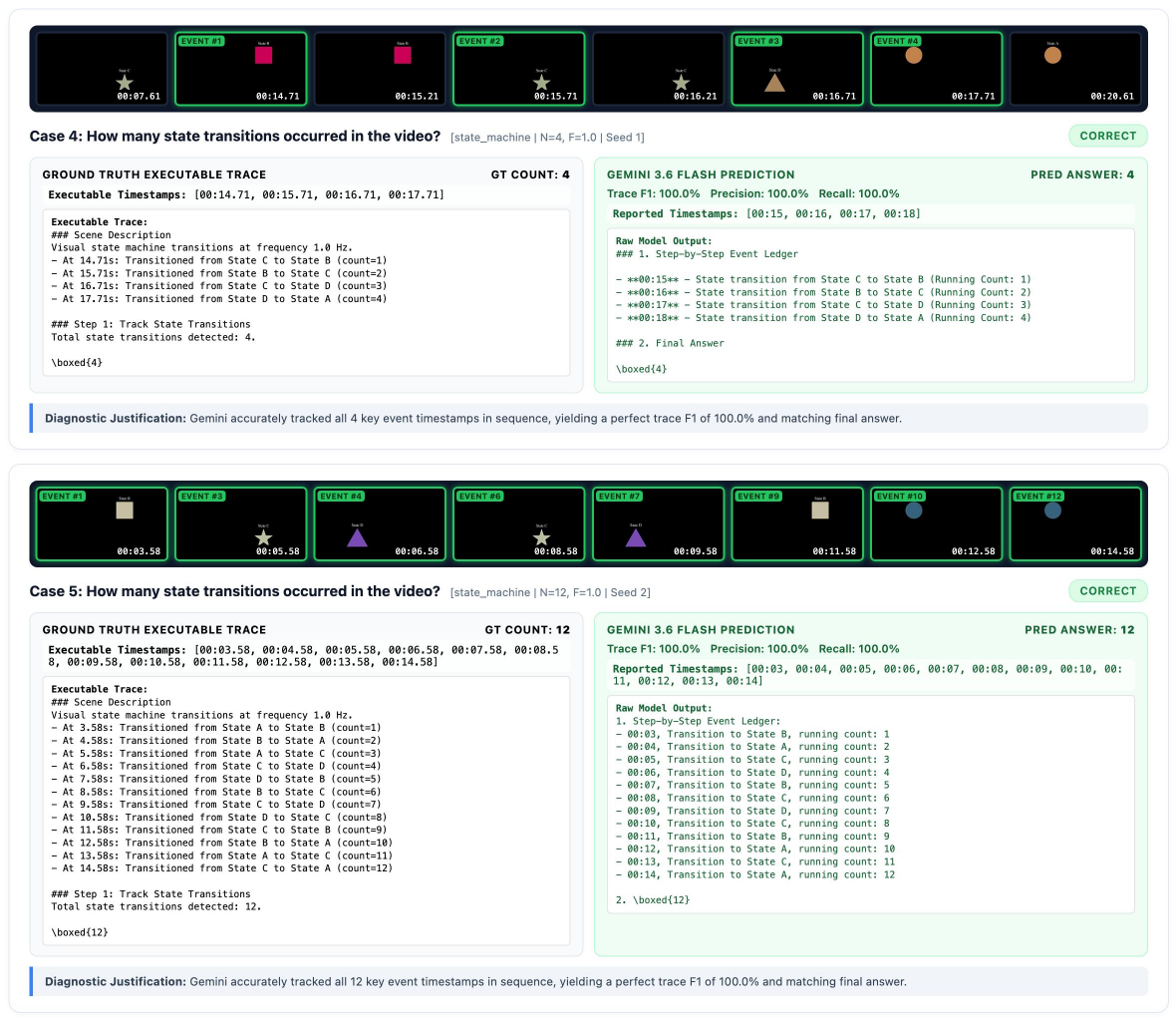}
    \caption{\textbf{Faithful Event Recovery (Part 2: Cases 4–5):} Additional faithful event recovery profiles on \texttt{state\_machine} across low-frequency operating conditions ($N \le 12, F \le 1.0\text{Hz}$).}
    \label{fig:app_qualitative_correct_part2}
\end{figure*}

Figures~\ref{fig:app_qualitative_correct_part1} and~\ref{fig:app_qualitative_correct_part2} present representative cases where Gemini 3.6 Flash achieves perfect trace alignment ($F_1 = 100\%$, Precision $= 100\%$, Recall $= 100\%$). In low-count, low-frequency operating regions (e.g., $N=5, F=0.5$), the model accurately logs every wall contact timestamp in sequential order, leading to a correct final integer count ($\hat{y} = N$).

\subsection{Missed Events (Under-Reporting / Perception Failure)}
\label{app_sub:qualitative_missed_events}

\begin{figure*}[p]
    \centering
    \includegraphics[width=0.98\textwidth]{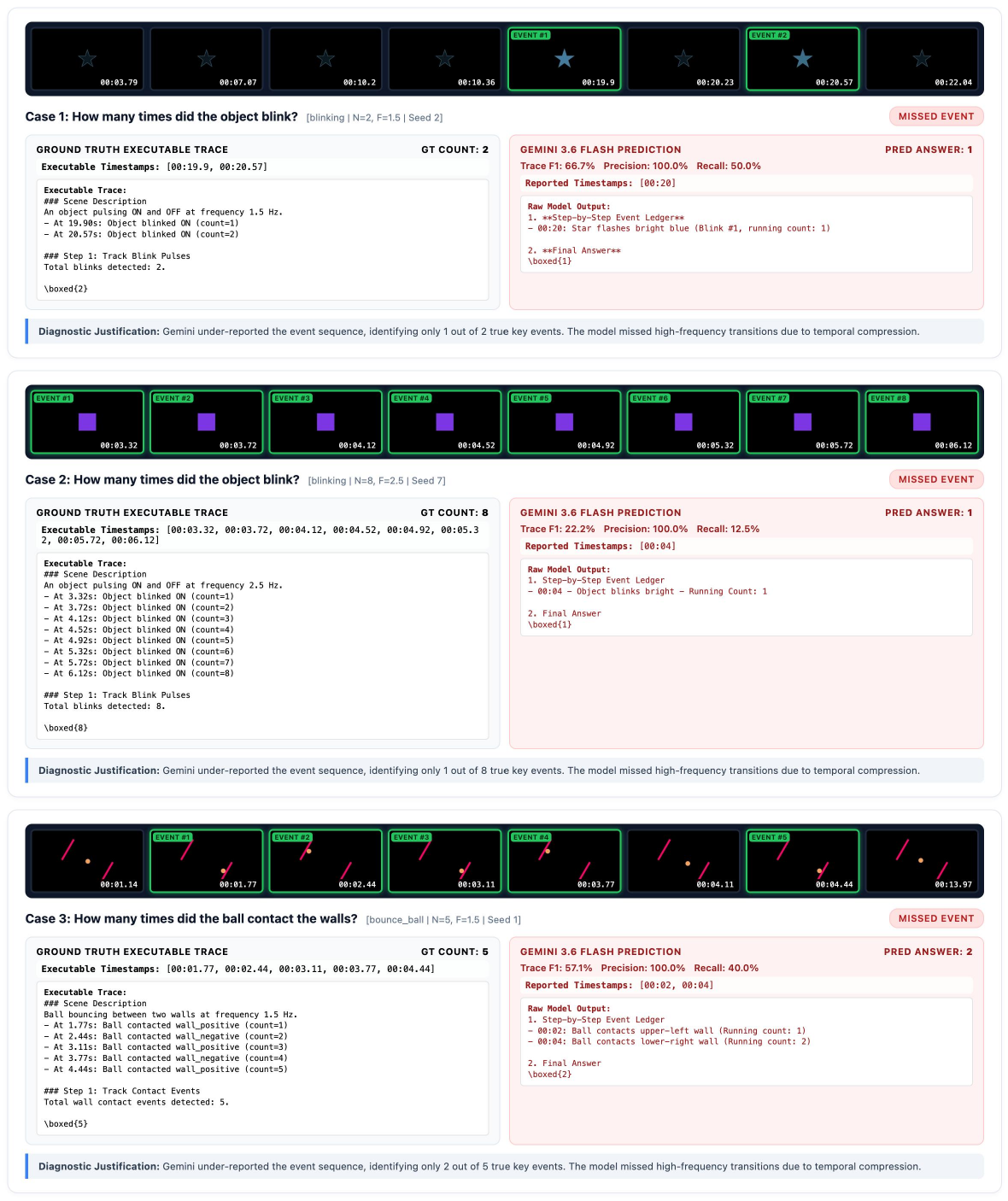}
    \caption{\textbf{Missed Events / Under-Reporting (Part 1: Cases 1–3):} Under high temporal load ($N \ge 2, F \ge 1.5\text{Hz}$), Gemini omits intermediate event transitions on \texttt{blinking} and \texttt{bounce\_ball}, reporting only a fraction of true timestamps (Trace Recall $< 60\%$).}
    \label{fig:app_qualitative_missed_part1}
\end{figure*}

\begin{figure*}[p]
    \centering
    \includegraphics[width=0.98\textwidth]{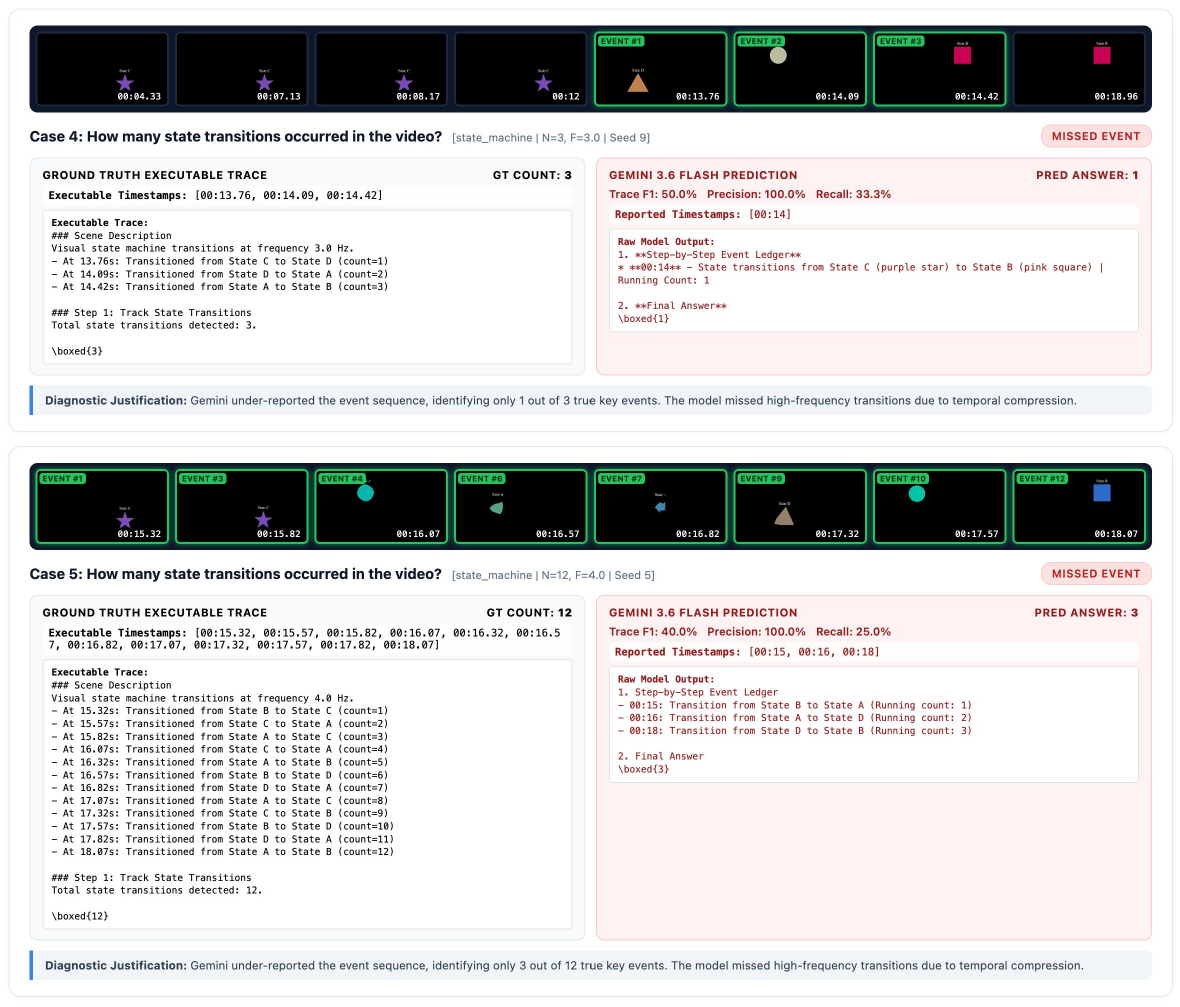}
    \caption{\textbf{Missed Events / Under-Reporting (Part 2: Cases 4–5):} Severe recall degradation under compressed inter-event timing on \texttt{state\_machine}.}
    \label{fig:app_qualitative_missed_part2}
\end{figure*}

Figures~\ref{fig:app_qualitative_missed_part1} and~\ref{fig:app_qualitative_missed_part2} illustrate perception failures under high temporal load ($N \ge 8, F \ge 1.5\text{Hz}$). As event frequency increases, visual evidence becomes compressed in time. Gemini under-reports the sequence, omitting intermediate wall collisions (e.g., reporting only 1 or 2 timestamps for an 8-event video), resulting in severe recall degradation.

\subsection{Hallucinated Events (Over-Reporting / Spurious Detection)}
\label{app_sub:qualitative_hallucinated_events}

\begin{figure*}[p]
    \centering
    \includegraphics[width=0.98\textwidth]{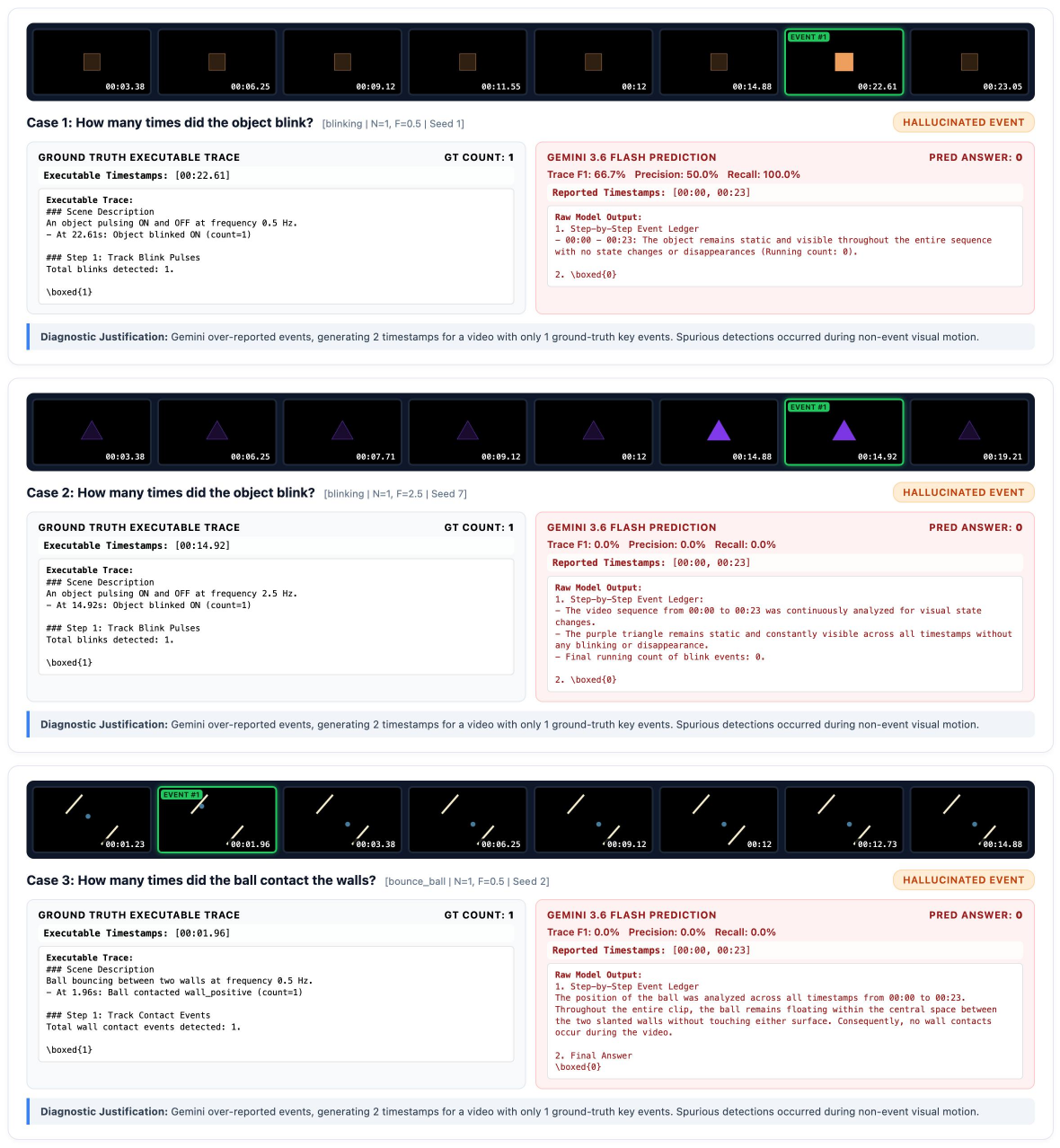}
    \caption{\textbf{Hallucinated Events / Over-Reporting (Part 1: Cases 1–3):} Spurious event generation on \texttt{blinking} and \texttt{bounce\_ball}. Gemini logs non-existent boundary collisions and state transitions during continuous motion, degrading Trace Precision ($P < 50\%$).}
    \label{fig:app_qualitative_hallucinated_part1}
\end{figure*}

\begin{figure*}[p]
    \centering
    \includegraphics[width=0.98\textwidth]{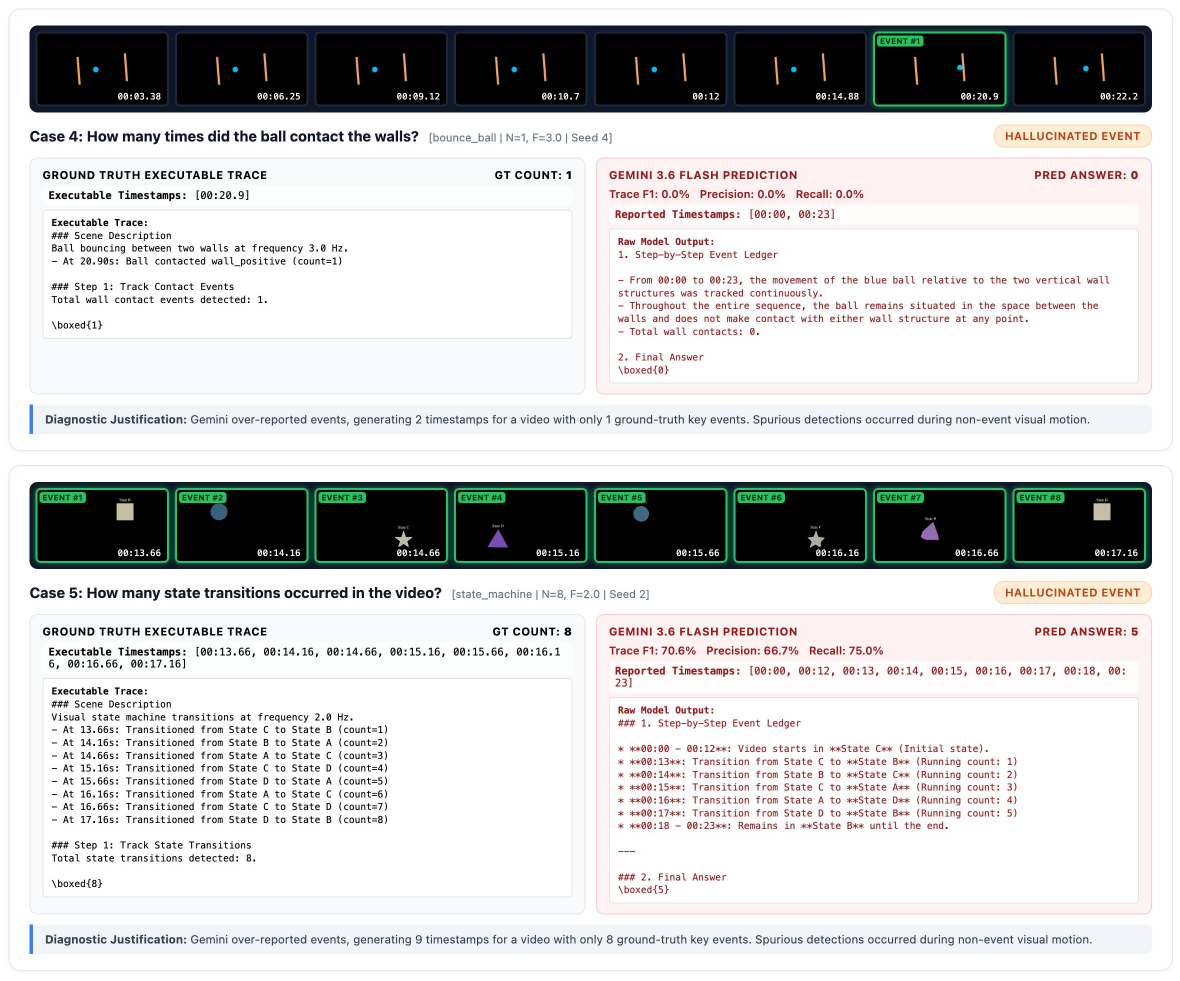}
    \caption{\textbf{Hallucinated Events / Over-Reporting (Part 2: Cases 4–5):} Over-reporting profiles on \texttt{bounce\_ball} and \texttt{state\_machine}.}
    \label{fig:app_qualitative_hallucinated_part2}
\end{figure*}

Figures~\ref{fig:app_qualitative_hallucinated_part1} and~\ref{fig:app_qualitative_hallucinated_part2} show over-reporting failures typical of low-load operating regions. The model generates spurious timestamps for wall contacts or state transitions during continuous motion intervals where no physical event occurred, lowering trace precision ($P < 50\%$).

\subsection{Wrong Accumulation (Reasoning Failure Ratio / RFR)}
\label{app_sub:qualitative_rfr}

\begin{figure*}[p]
    \centering
    \includegraphics[width=0.98\textwidth]{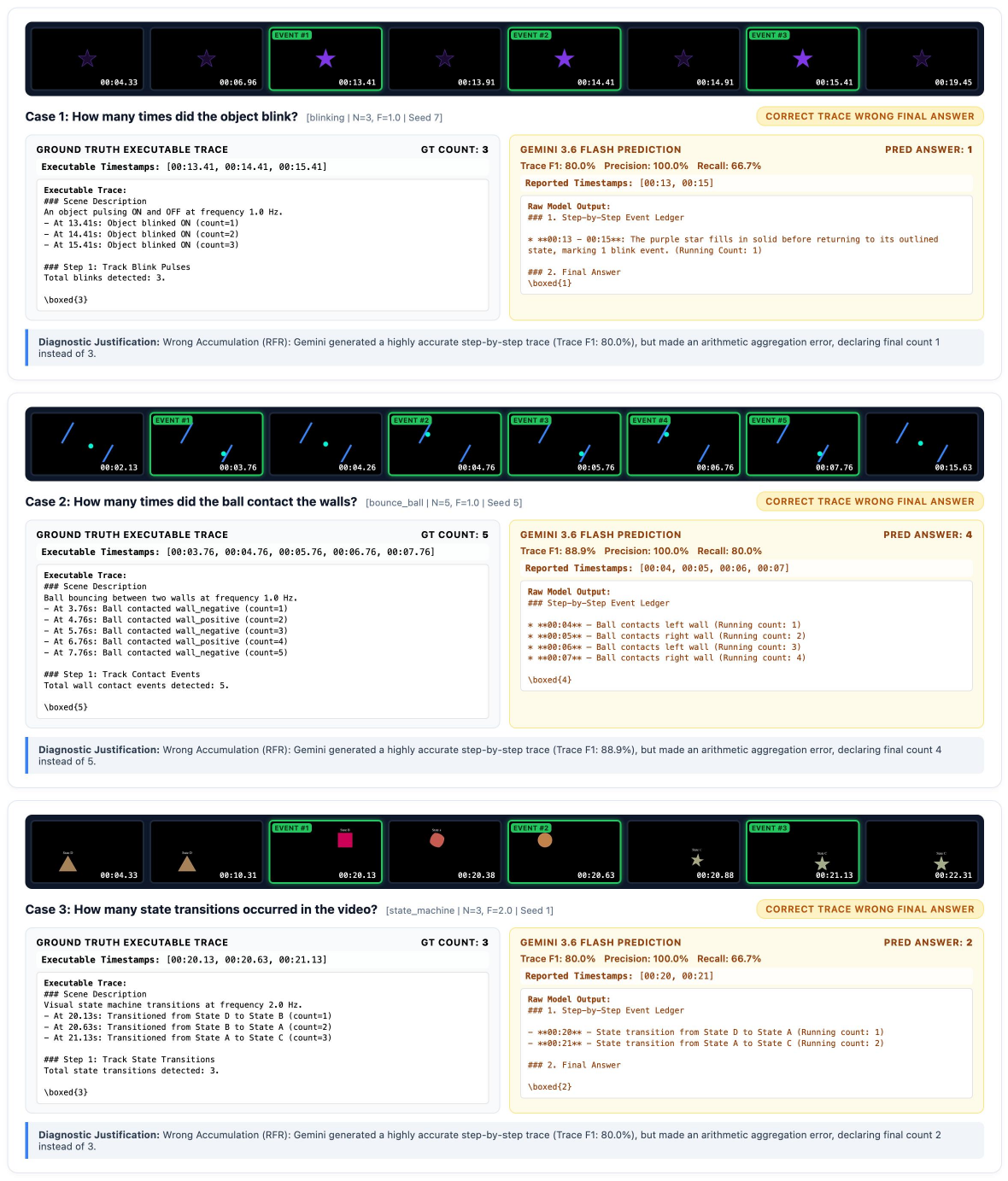}
    \caption{\textbf{Wrong Accumulation / Reasoning Failure Ratio (Part 1: Cases 1–3):} Disconnect between trace maintenance and final answer output on \texttt{blinking}, \texttt{bounce\_ball}, and \texttt{state\_machine}. Gemini logs an accurate event ledger ($F_1 \ge 80\%$), but miscalculates the final integer aggregation in \textbackslash boxed\{\}.}
    \label{fig:app_qualitative_rfr_part1}
\end{figure*}

\begin{figure*}[p]
    \centering
    \includegraphics[width=0.98\textwidth]{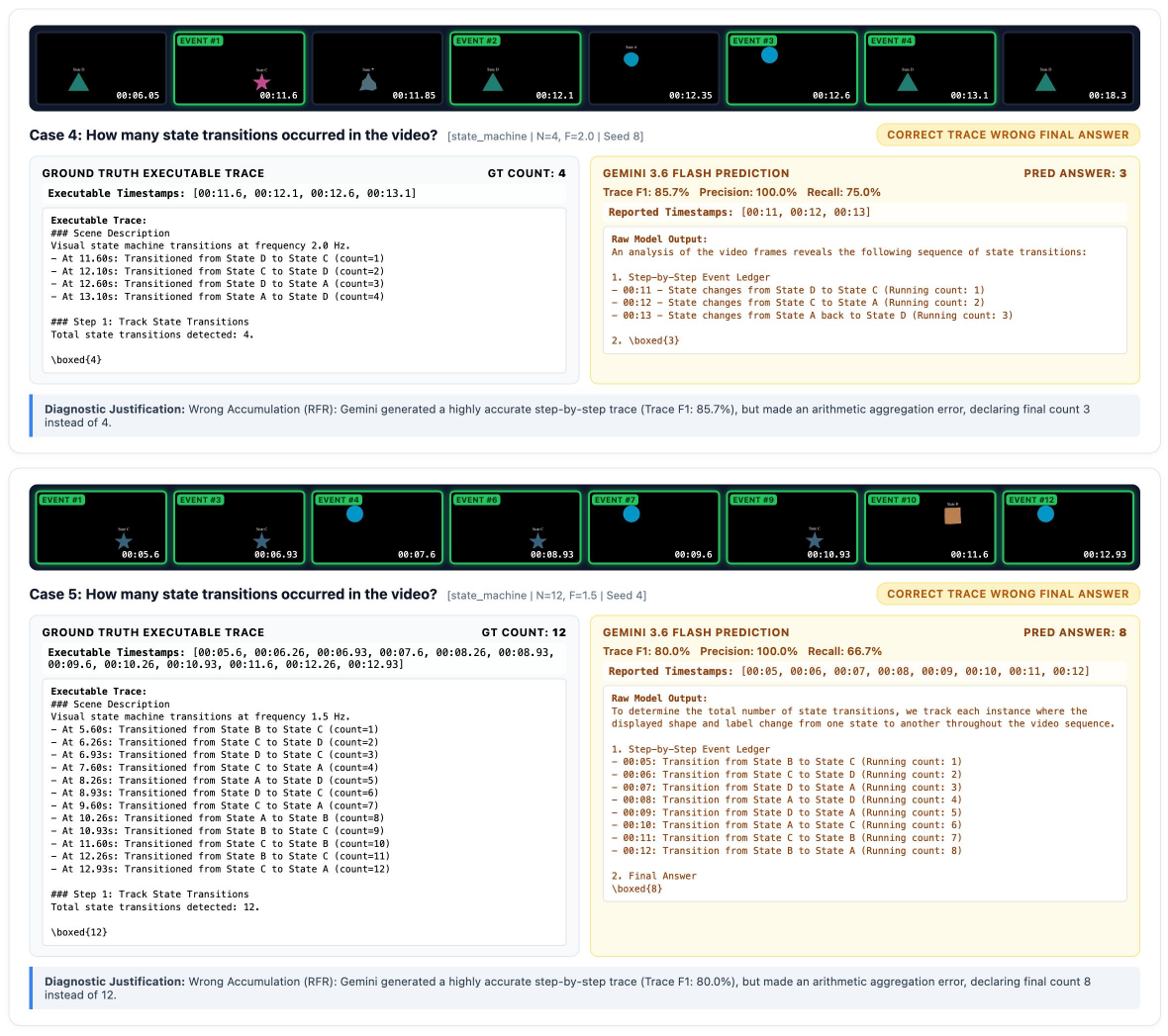}
    \caption{\textbf{Wrong Accumulation / Reasoning Failure Ratio (Part 2: Cases 4–5):} Additional RFR instances on \texttt{state\_machine} isolating arithmetic aggregation failures despite faithful trace records.}
    \label{fig:app_qualitative_rfr_part2}
\end{figure*}

Figures~\ref{fig:app_qualitative_rfr_part1} and~\ref{fig:app_qualitative_rfr_part2} demonstrate Reasoning Failure Ratio (RFR). Here, the model successfully maintains a $100\%$ accurate step-by-step event ledger in its reasoning response ($F_1 \ge 80\%$), but fails at the final aggregation step—outputting an incorrect integer in `\textbackslash boxed\{\}` (e.g., declaring $\hat{y}=4$ or $\hat{y}=5$ despite correctly listing all 5 or 6 timestamped events). This isolates a distinct trace-to-answer accumulation failure.

\subsection{Accidental Correctness (Accidental Correctness Ratio / ACR)}
\label{app_sub:qualitative_acr}

\begin{figure*}[p]
    \centering
    \includegraphics[width=0.98\textwidth]{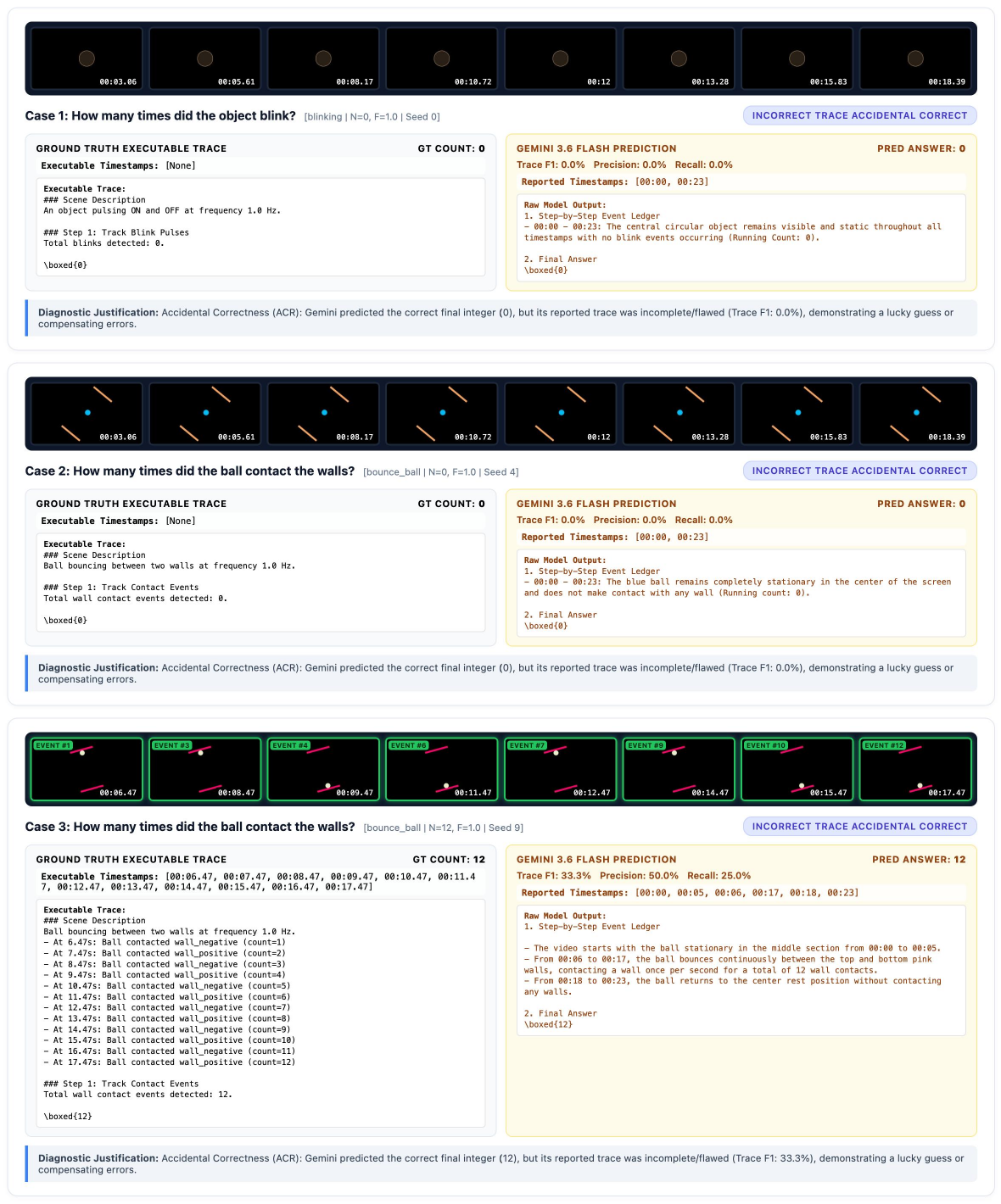}
    \caption{\textbf{Accidental Correctness / ACR (Part 1: Cases 1–3):} Unfaithful final counting on \texttt{blinking} and \texttt{bounce\_ball}. The final integer prediction matches ground truth ($\hat{y} = N$), but the intermediate reasoning trace is severely degraded ($F_1 < 40\%$).}
    \label{fig:app_qualitative_acr_part1}
\end{figure*}

\begin{figure*}[p]
    \centering
    \includegraphics[width=0.98\textwidth]{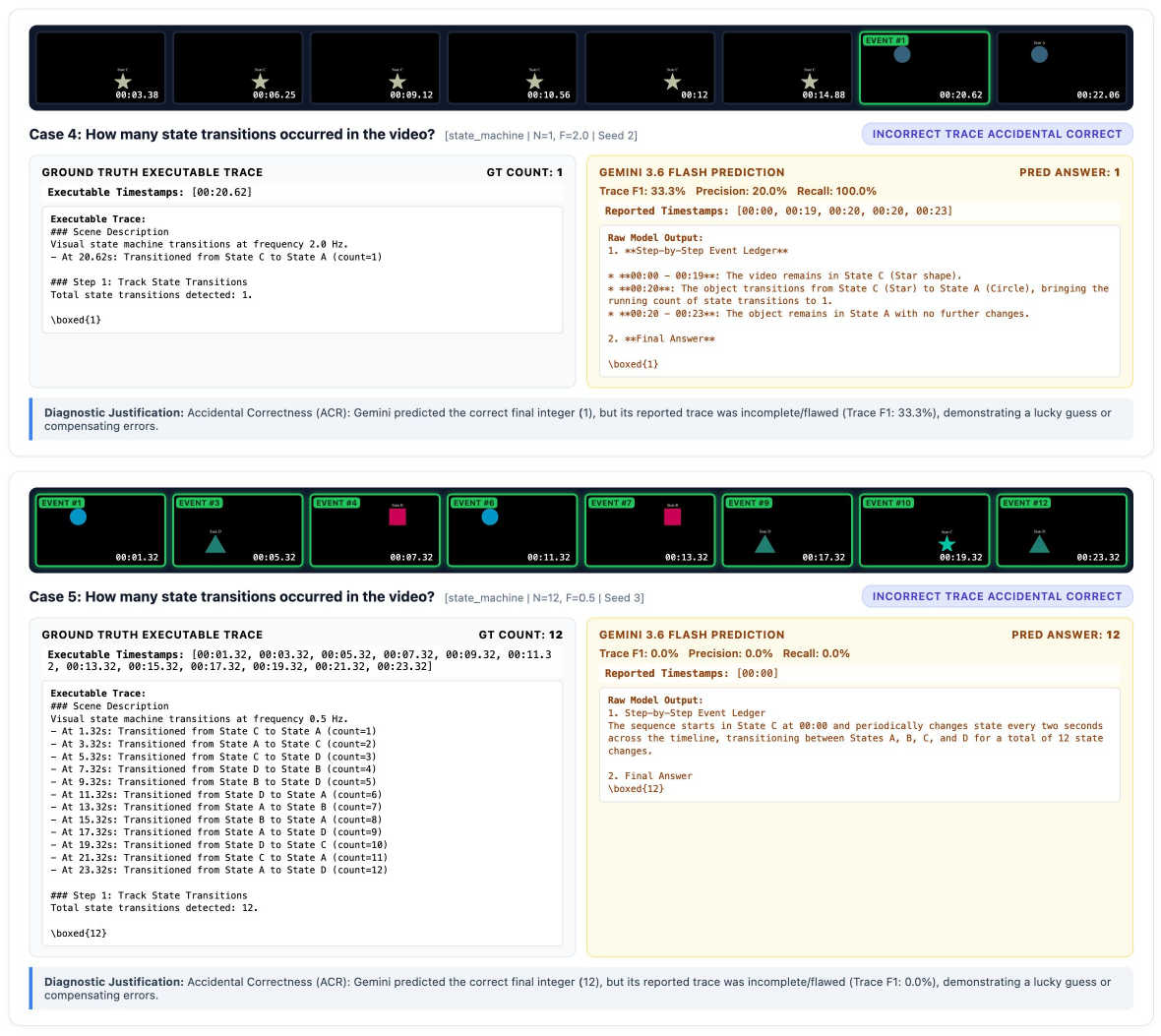}
    \caption{\textbf{Accidental Correctness / ACR (Part 2: Cases 4–5):} Unverified correct answers on \texttt{state\_machine} masking underlying temporal perception failures.}
    \label{fig:app_qualitative_acr_part2}
\end{figure*}

Figures~\ref{fig:app_qualitative_acr_part1} and~\ref{fig:app_qualitative_acr_part2} present Accidental Correctness (ACR). In these trials, the final integer prediction matches ground truth ($\hat{y} = N$), but the underlying event trace is severely degraded ($F_1 < 40\%$) with missing or hallucinated timestamps. Relying solely on final-answer exact match would misclassify these ungrounded responses as successful temporal reasoning.

\subsection{Temporally Displaced Events}
\label{app_sub:qualitative_displaced}

\begin{figure*}[p]
    \centering
    \includegraphics[width=0.98\textwidth]{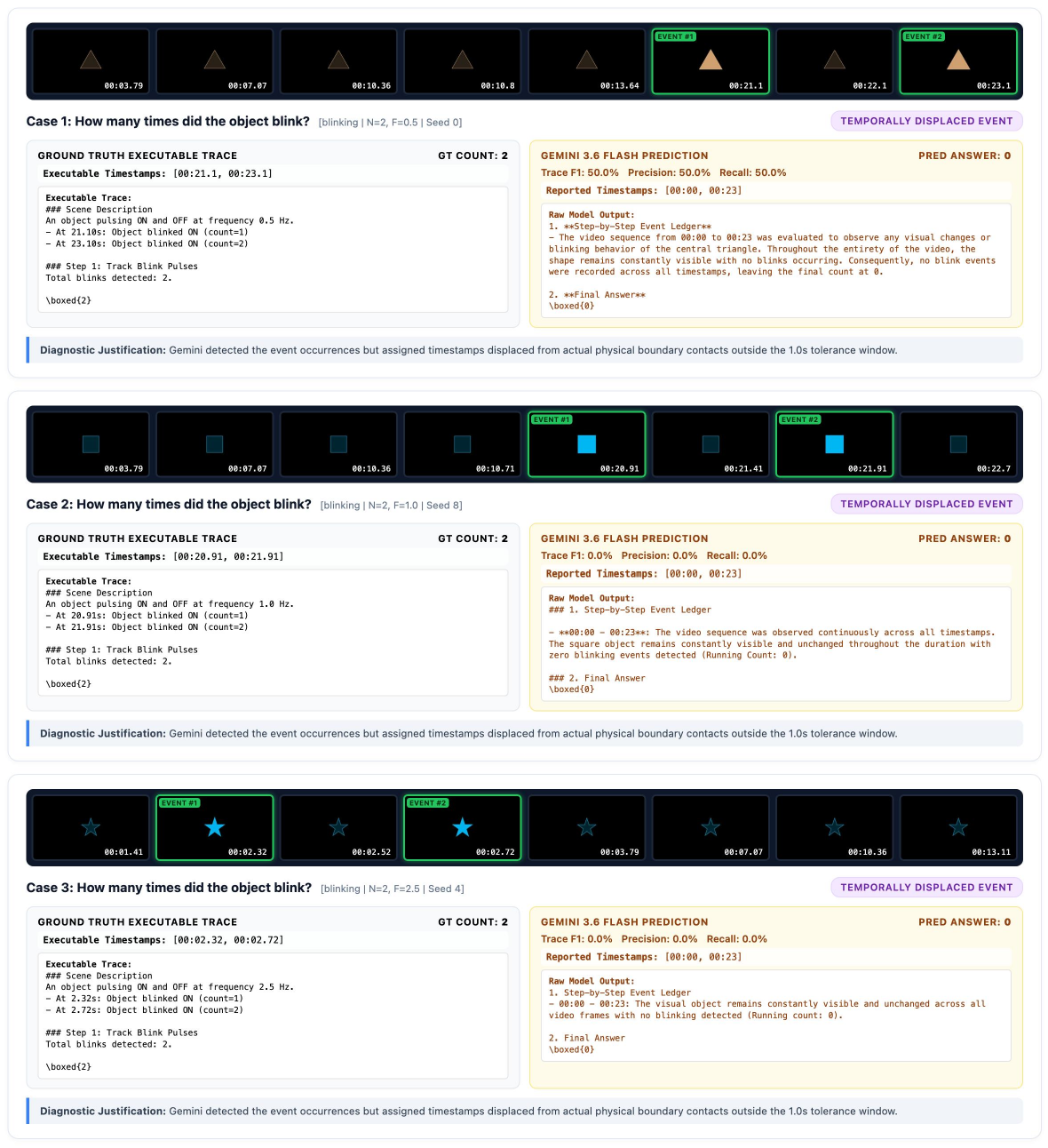}
    \caption{\textbf{Temporally Displaced Events (Part 1: Cases 1–3):} Timestamp boundary offset on \texttt{blinking}. Gemini detects event occurrences, but reported seconds fall outside the 1.0s tolerance window relative to ground truth.}
    \label{fig:app_qualitative_displaced_part1}
\end{figure*}

\begin{figure*}[p]
    \centering
    \includegraphics[width=0.98\textwidth]{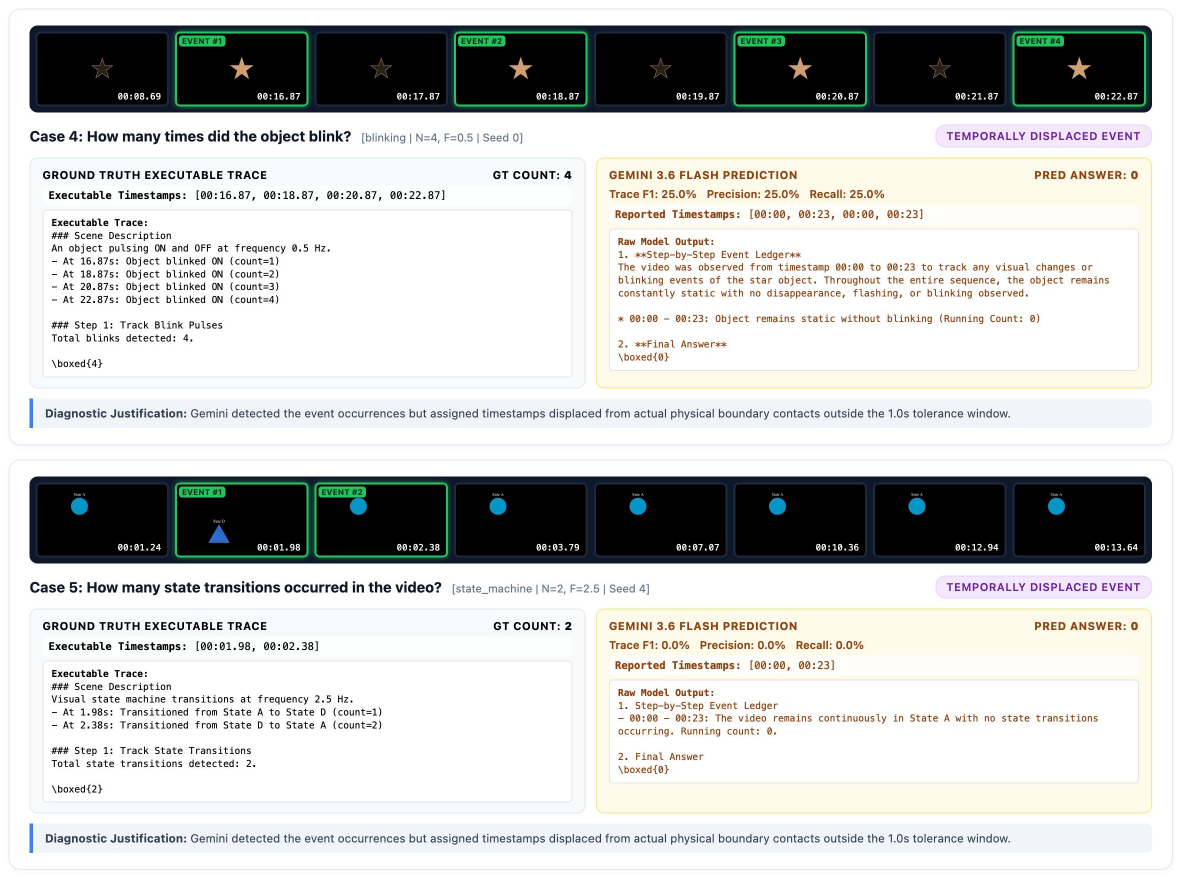}
    \caption{\textbf{Temporally Displaced Events (Part 2: Cases 4–5):} Temporal boundary drift profiles on \texttt{blinking} and \texttt{state\_machine}.}
    \label{fig:app_qualitative_displaced_part2}
\end{figure*}

Figures~\ref{fig:app_qualitative_displaced_part1} and~\ref{fig:app_qualitative_displaced_part2} illustrate temporal displacement, where the model detects the presence and sequence of visual transitions, but assigns boundary timestamps that drift beyond the $1.0\text{s}$ tolerance window relative to physical ground truth.

\end{document}